\RequirePackage{fix-cm}
\documentclass[smallcondensed]{svjour3}     
\smartqed  
\usepackage{graphicx}
\usepackage{caption}
\usepackage{color, colortbl}
\definecolor{Gray}{gray}{0.9}
\definecolor{GrayL}{gray}{0.95}
\usepackage{multirow}
\usepackage[inline]{enumitem}
\usepackage{amsmath}
\usepackage{bbm}
\usepackage{natbib}
\usepackage{hyperref}       
\usepackage{url}            
\usepackage[ruled,vlined]{algorithm2e}

\DeclareMathOperator*{\argmin}{arg\,min}
\DeclareMathOperator*{\argmax}{arg\,max}
\setlist[itemize]{leftmargin=3mm}

\begin{document}

\title{Using Error Decay Prediction to Overcome Practical Issues of Deep Active Learning for Named Entity Recognition}

\titlerunning{Overcoming Practical Issues of Deep Active Learning}        

\author{Haw-Shiuan Chang \and Shankar Vembu \and Sunil Mohan \and Rheeya Uppaal \and \\  Andrew McCallum
}


\institute{Haw-Shiuan Chang (work done while interning at CZI) \and Rheeya Uppaal \and Andrew McCallum  \at
              University of Massachusetts Amherst, College of Information and Computer Science, Amherst, MA, USA \\
             \email{\{hschang,ruppaal,mccallum\}@cs.umass.edu}           
           \and
           Sunil Mohan \at
           Chan Zuckerberg Initiative (CZI), Redwood City, CA, USA \\
           \email{smohan@chanzuckerberg.com}
           \and
           Shankar Vembu \at
           Work done while SV was at CZI. \\
           \email{shankar@argmix.com}
      }


\maketitle

\begin{abstract}
Existing deep active learning algorithms achieve impressive sampling efficiency on natural language processing tasks. However, they exhibit several weaknesses in practice, including (a) inability to use uncertainty sampling with black-box models, (b) lack of robustness to labeling noise, and (c) lack of transparency. In response, we propose a transparent batch active sampling framework by estimating the error decay curves of multiple feature-defined subsets of the data.

Experiments on four named entity recognition (NER) tasks demonstrate that the proposed methods significantly outperform diversification-based methods for black-box NER taggers, and can make the sampling process more robust to labeling noise when combined with uncertainty-based methods. Furthermore, the analysis of experimental results sheds light on the weaknesses of different active sampling strategies, and when traditional uncertainty-based or diversification-based methods can be expected to work well.
\end{abstract}

\section{Introduction}

Deep neural networks achieve state-of-the-art results on many tasks, especially when a large amount of training data is available. Their success highlights the importance of reducing the cost of collecting labels on a large scale.
Active learning can be used to select data samples that will most benefit a predictor's training, thereby reducing the amount of labeled data needed without hurting the predictor's accuracy. The effectiveness of uncertainty and disagreement-based\footnote{In this work, we view disagreement scores as a kind of uncertainty estimation to simplify our discussion.} active learning methods have been demonstrated on several datasets for shallow predictors~\citep{settles2008analysis,settles2009active}, and more recently also for deep learning predictors~\citep{gal2017deep,shendeep,siddhant2018deep}. Nevertheless, random sampling is still the most popular method to build new datasets in several domains, including natural language processing ~\citep{tomanek2009web}. This is due to the practical issues of deploying uncertainty-based active sampling~\citep{settles2011theories,lowell2018transferable}, including its limited {\em applicability}, {\em robustness}, and {\em transparency}.

{\em Applicability:} Uncertainty sampling selects samples with the lowest prediction confidence of a predictor and collects their labels. However, the uncertainty in prediction may be hard to estimate for some complex models. For example, a relation extraction system is often a pipeline that includes named entity extraction, entity linking, and sentence classification. Combining the uncertainty of different components is difficult and usually relies on ad hoc approaches~\citep{reichart2008multi}. Another common case is that state-of-the-art commercial software packages or online services may only provide final predictions (i.e., a black-box predictor)~\citep{wang2017active}. Our first research question is whether a general active learning method can be based only on the predictions from such black-box predictors.

{\em Robustness:} Uncertainty sampling assumes that the labels of uncertain examples are the most informative training data. However, the strategy may select outliers or ambiguous examples for annotators due to their low confidence scores. This selection bias often introduces substantial labeling noise into the dataset~\citep{mussmann2018relationship}. In fact, outlier detection methods often improve prediction performance by adopting a sample selection strategy that is exactly opposite to uncertainty sampling~\citep{bouguelia2018agreeing}. It is usually hard to distinguish an informative training example from a misleading outlier, and blindly adopting uncertainty sampling may result in much worse performance than random sampling in practice. Our second research question is whether a general active learning method can be robust to labeling noise without relying on prior knowledge.





{\em Transparency:} Active sampling introduces a sampling bias, but we often know little about its effects on the performance in different aspects. This lack of transparency causes many practical issues. Not understanding why an active learning method works well for a dataset, we have few insights into whether the sampling method will work for another similar dataset well enough to compensate the effort of trying active learning, whether the deep predictors will perform worse in some crucial aspects/classes, whether the selection will emphasize undesirable biases (e.g., a person's race), and whether the sampling efficiency improvement will vanish in the long run or after switching the deep predictor for uncertainty estimation.

Furthermore, practitioners need to choose a sampling method for a new task and a given deep predictor before collecting labels (e.g., random, uncertainty or diversification sampling, or combination of the above). Existing deep active learning studies usually focus on improving prediction performance without exploring why and how much the performance gains depend on the predictor's specific ways of modeling feature interactions. The lack of such insights makes the choice extremely difficult~\citep{lowell2018transferable}. Thus, our third research question is whether we can have a general analysis tool to provide insights into when and why an active sampling method works, and forecast the potential benefits in different aspects based on a small number of existing labels. 

\begin{figure*}[!t]
\includegraphics[width=\linewidth]{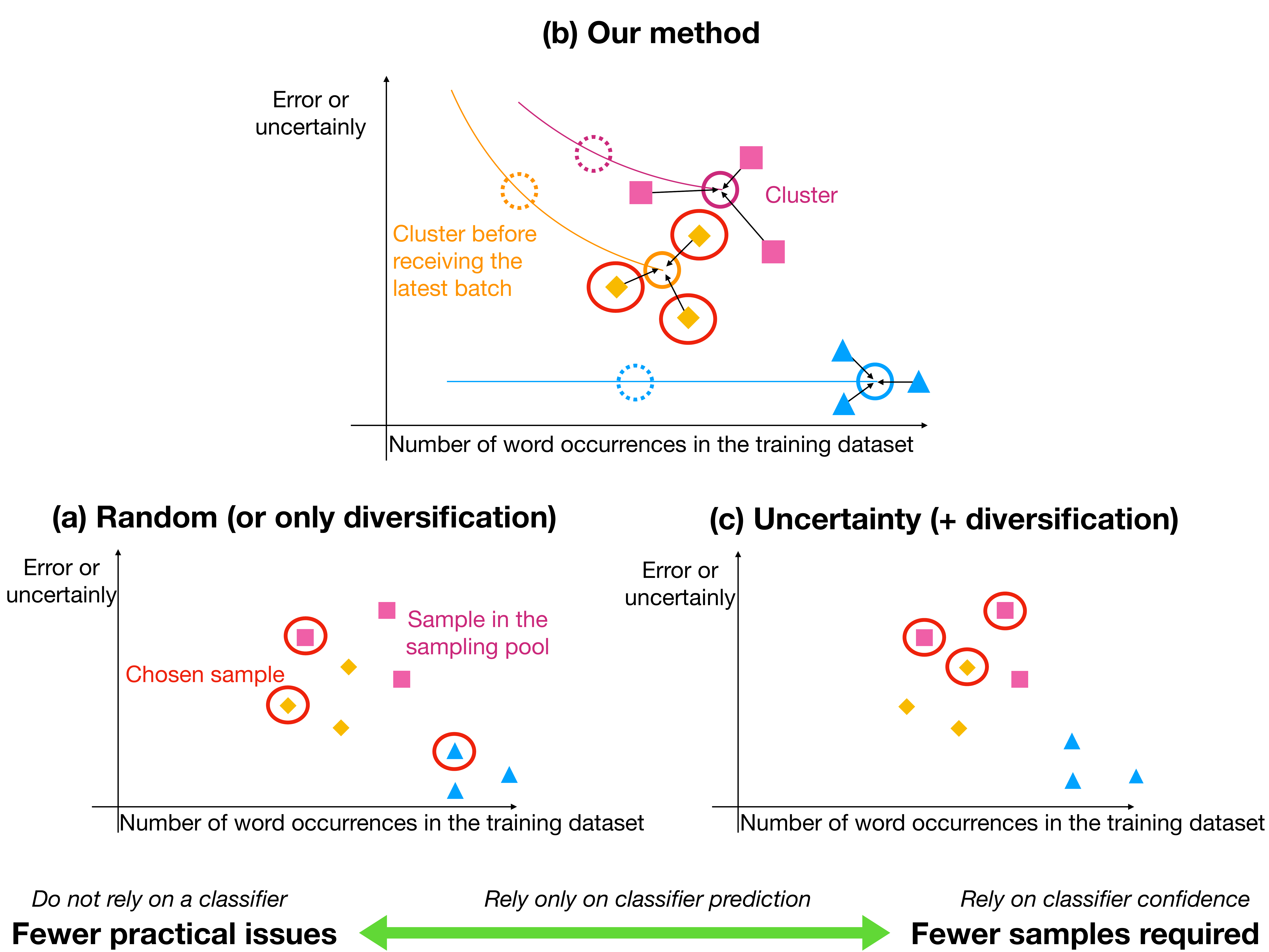}
\caption{Comparison of sampling strategies. Each node is a word in an NER problem and we plot its prediction error/uncertainty versus its count in the current training dataset. Random or diversification-based sampling (a) often selects samples irrelevant to the task, such as a blue triangle node. Uncertainty sampling (c) prefers to select samples with the highest error/uncertainty (e.g., pink square nodes). However, their error decay could be small because of some inherent label noise. Our method (b) is a novel balance in this spectrum which usually provides better performance than diversification while being more robust, explainable, and applicable than uncertainty sampling (+ diversification).
}
\label{fig:illustration}
\end{figure*}

To answer the above research questions, we first illustrate three kinds of samples in Figure~\ref{fig:illustration}. The blue triangle nodes are easy samples, which means their prediction errors are low and remain almost unchanged if we collect more such labels. The pink square nodes represent samples with noisy labels, which means their prediction errors or uncertainties are high but decrease slowly as more labels are seen. To select informative samples like the orange diamond nodes, we propose a batch active learning framework that maximizes the error reduction of each sample batch.

\begin{figure}[!t]
  \centering
\includegraphics[width=0.7\linewidth]{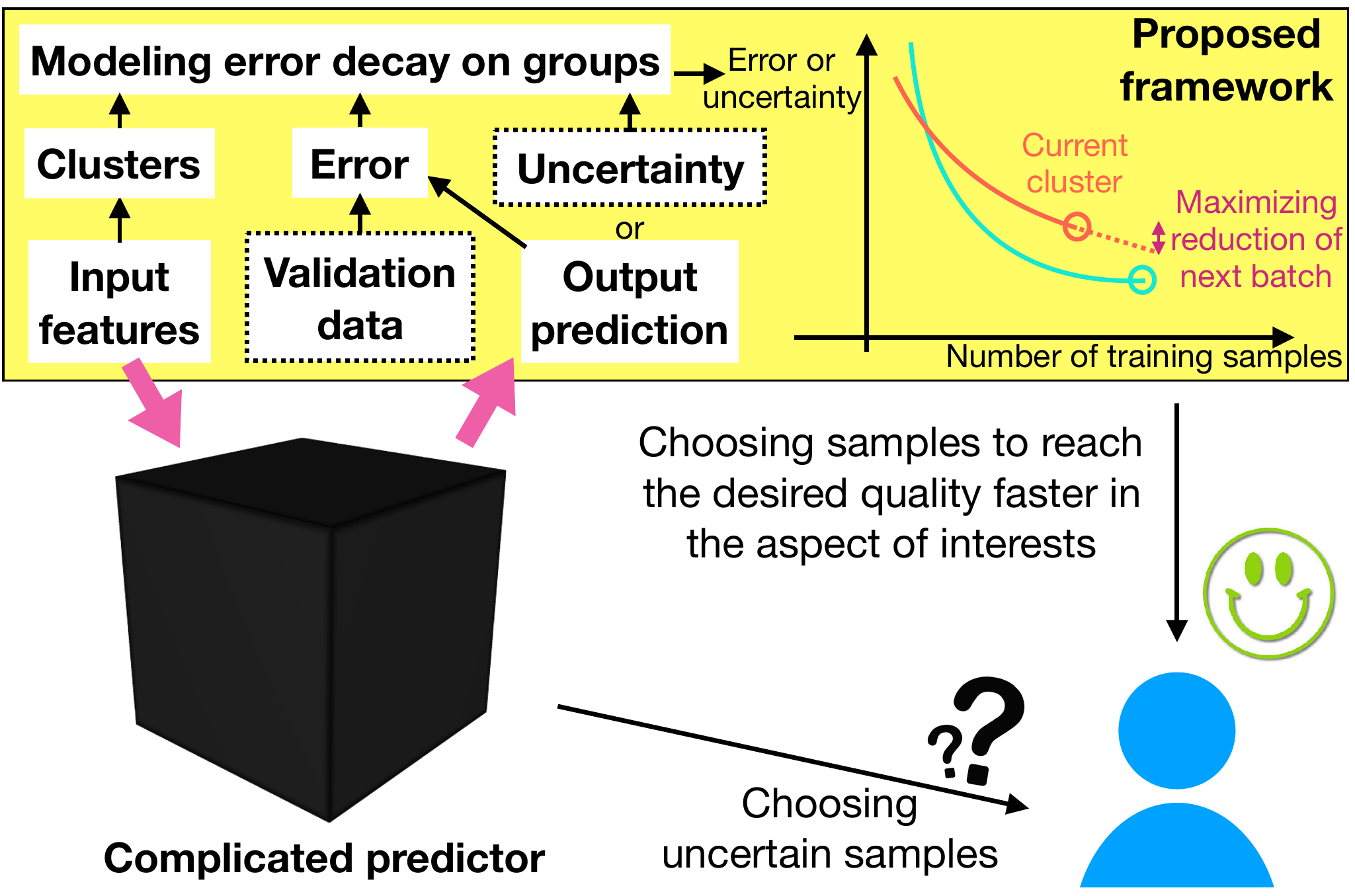}
\captionof{figure}{The uncertainty sampling for a black-box predictor is hard to interpret. Our proposed framework predicts the reduction of error/uncertainty on clusters of samples, optionally with the help of validation data. The resulting strategies provide a more direct explanation of the choices made by batch active learning.
}
\label{fig:AAAI_first_page}
\end{figure}

In this framework, we propose three sampling methods. When it's not possible to access the uncertainty of the deep predictor but labels of a validation dataset are available, we cluster the samples and predict the validation error reduction on each cluster as shown in Figure~\ref{fig:AAAI_first_page}. The first method is called error decay on groups (EDG). Without a validation dataset, the second method approximates the error decay using prediction changes on groups. Finally, if uncertainty is available, we model the uncertainty reduction of each sample. In Section\ref{sec:method}, we describe the first method in detail, and view the second and third methods as its extensions and denote them as EDG\_ext1 and EDG\_ext2, respectively.



Analyzing error decay on clusters results in several practical benefits. We can model the error decay using only model predictions; noisy samples will not be selected due to their low error decay; the strategy of our active sampling is interpretable (e.g., sampling more uppercased words in a named entity recognition problem due to their larger prediction error decay); and we can predict the performance gains from active learning in different clusters/aspects.
Nevertheless, having only a few labels, we found that achieving the practical advantages sacrifices some sampling efficiency. To feasibly approximate the error reduction, we assume independence between samples in different clusters. For instance, if a sample is a word in a sentence, our selection method assumes its neighboring words do not affect the error decay of the word. In practice, violation of this assumption causes sub-optimal sampling efficiency when compared to uncertainty-based active learning.


We apply our framework to named entity recognition (NER) problems where our independence assumption is famously violated, and comprehensively evaluate the pros and cons of the proposed framework.\footnote{The proposed methods could be extended to other types of classification and sequential tagging problems given proper ways of clustering samples and a family of proper error decay functions.}
We summarize our results from NER experiments in Table~\ref{tb:result_sum}, which shows that our framework provides novel ways to trading partial sampling efficiency gain for better {\em applicability}, {\em robustness}, and {\em transparency}.
\begin{table}[!t]
  \centering
  \scalebox{1}{
  \begin{tabular}[b]{|c|ccc|cc|} \hline
   & \multicolumn{3}{|c|}{Proposed} & \multicolumn{2}{|c|}{Classic} \\
   & EDG & \_ext1 & \_ext2 & US (+ Div) & Rnd/Div \\
   \hline
   Interpretable & \cellcolor{Gray} & \cellcolor{Gray} & \multirow{2}{*}{No} & \multirow{2}{*}{No} & \cellcolor{Gray} \\
   Strategy & \cellcolor{Gray}\multirow{-2}{*}{Yes} & \cellcolor{Gray}\multirow{-2}{*}{Yes} & & & \cellcolor{Gray}\multirow{-2}{*}{Yes}\\ \hline
   Robustness & \cellcolor{Gray} & \cellcolor{GrayL} & \cellcolor{GrayL} & \multirow{2}{*}{Low} & \cellcolor{Gray} \\
   to Noise & \cellcolor{Gray}\multirow{-2}{*}{High} & \cellcolor{GrayL}\multirow{-2}{*}{Med} & \cellcolor{GrayL}\multirow{-2}{*}{Med} & & \cellcolor{Gray}\multirow{-2}{*}{High}\\ \hline
   Output & \cellcolor{GrayL} & \cellcolor{GrayL} & \multirow{2}{*}{Prob} & \multirow{2}{*}{Prob} & \cellcolor{Gray}  \\ 
   Required & \cellcolor{GrayL}\multirow{-2}{*}{Pred} & \cellcolor{GrayL}\multirow{-2}{*}{Pred} & & & \cellcolor{Gray}\multirow{-2}{*}{None}\\ \hline
   Validation & \multirow{2}{*}{Yes} & \cellcolor{Gray} & \cellcolor{Gray} & \cellcolor{Gray} & \cellcolor{Gray}  \\ 
   Required & & \cellcolor{Gray}\multirow{-2}{*}{No} & \cellcolor{Gray}\multirow{-2}{*}{No} & \cellcolor{Gray}\multirow{-2}{*}{No} & \cellcolor{Gray}\multirow{-2}{*}{No}\\ \hline
   Label Cost& \cellcolor{GrayL} & \cellcolor{GrayL}  & \cellcolor{Gray} & \cellcolor{Gray}  & \multirow{2}{*}{Low} \\ 
    Reduction & \cellcolor{GrayL}\multirow{-2}{*}{Med} & \cellcolor{GrayL}\multirow{-2}{*}{Med} & \cellcolor{Gray}\multirow{-2}{*}{High} & \cellcolor{Gray}\multirow{-2}{*}{High} &\\
   \hline
  
  \end{tabular}
  }
  \captionof{table}{Comparison of different sampling approaches. We compare transparency (i.e., interpretable strategy), robustness to labeling noise, required output from the predictor, whether validation data is required or not, and cost reduction of collecting clean labels.  US means uncertainty sampling (including disagreement-based sampling), Rnd means random sampling, and Div means diversification-based sampling. Med, pred, and prob means medium, predictor prediction, and prediction probability, respectively. In black-box models, only predictions of the predictor are available. A more ideal property is highlighted with a darker shade in the table.}
  \label{tb:result_sum}
\end{table}

\subsection{Summary of Main Contributions}
\label{sec:contribution}
\setlist{nolistsep}
\begin{enumerate}
\item We propose a novel active learning method, EDG, which models validation error decay curves on clusters of samples.  We extend the method by replacing validation error decay with prediction difference decay or uncertainty decay to avoid relying on validation data. This demonstrates the flexibility of EDG framework.
\item In one synthetic and three real-world NER datasets, we show that EDG significantly outperforms the diversification baseline for the black-box model with or without the help of validation data. This demonstrates the effectiveness and applicability of EDG framework.
\item Modeling the error decay on clusters can be used as an analysis tool for arbitrary active learning methods. The experimental results show that no single method always wins and the proposed analysis tool provides intuitions and guidelines on selecting a specific method. This demonstrates the transparency of EDG framework.
\item We propose a new evaluation method based on pseudo labels. The method allows us to test active learning methods on a large sampling pool and test their robustness to systematic labeling noise.
\item In the experiments on pseudo labels and synthetic dataset, we show that combining EDG with state-of-the-art uncertainty sampling methods (i.e., choosing the samples with the highest uncertainty decay rather than uncertainty) improves sampling efficiency in the presence of systematic labeling noise, random labeling noise, and model change. This demonstrates the robustness of EDG framework.
\end{enumerate}

\section{Related Work}

Several studies aim at making active learning practical~\citep{settles2011theories}. For example, \citet{phillips2018interpretable} extend the techniques of interpreting a classifier's predictions to explain the active sampling process, but still rely on the uncertainty estimation of the classifier. \citet{bloodgood2009method} survey and propose stopping criteria based on overall error decay. Instead, we propose a novel active sampling method by modeling error decay on groups of samples.

Recently, active learning on black-box models has attracted research attention due to practical needs. \citet{wang2017active} focus on improving a black-box semantic role labeling model using another neural network with low transparency.  \citet{rubens2011active} propose estimating the variance of predictions of a black-box regressor, which is computationally prohibitive unless applied to simple models such as a linear regressor. 

Popular active learning methods such as uncertainty sampling are not robust to noise~\citep{mussmann2018relationship}, and there is a body of research in active learning (referenced below) to address this issue. However, the main focus of previous studies is to identify and post-process the noisy labels given some prior knowledge of the noise-to-signal ratio. Instead, we propose active learning methods which are inherently robust to noise by avoiding the selection of difficult samples for annotators in the first place. 

The methods to identify noisy labels are similar to the methods that measure the uncertainty of each sample~\citep{sheng2008get,bouguelia2018agreeing}, which could be based on disagreement between workers~\citep{sheng2008get,zhao2011incremental}, disagreement between manually created labels and automatically generated predictions ~\citep{bouguelia2015stream,khetan2017learning}, model uncertainty~\citep{sheng2008get,kremer2018robust,bouguelia2018agreeing}, or estimation of workers' quality~\citep{zhang2015active,khetan2017learning}. After identifying noisy labels, different methods adopt different strategies such as relabeling~\citep{sheng2008get,bouguelia2015stream}, excluding/down-weighting the noisy labels~\citep{khetan2017learning}, or both~\citep{zhao2011incremental,zhang2015active,bouguelia2018agreeing}, or acquiring high-quality labels~\citep{kremer2018robust}. 

Some approaches, such as~\citet{dasgupta2011two}, cluster input features and diversify samples by choosing them from different clusters, without considering information from the predictors being trained, and often showing only limited improvements in sampling efficiency. Recent approaches~\citep{settles2008analysis,wei2015submodularity,sener2017active,ravi2018meta} have combined uncertainty and diversification (e.g., by multiplying the informativeness and representativeness scores). By directly modeling the error reduction, our proposed methods naturally balance the two criteria without relying on model uncertainty estimation. This makes our methods robust to labeling noise and applicable to black-box models. 

Another direction for deep active learning is to learn an error reduction predictor or a sample selector. However, the selection model is either only applicable to a simple model like na\"ive Bayes~\citep{roy2001toward,fu2018scalable}, or requires a large amount of data to train a complex predictor or selector. The training data for the non-transparent selection model usually needs to come from a similar task. For instance, they could be images with different labels for image classification, other users for recommendation, or another language for NER~\citep{bachman2017learning,fang2017learning,ravi2018meta}. Although transferring the error reduction predictor between different types of datasets is possible~\citep{KonyushkovaSF17}, it is unclear on which dataset pairs such a transfer would work~\citep{koshorek2019limits}. 

A challenge related to active learning is to discover blind spots of the predictor (also called unknown unknowns). \citet{lakkaraju2017identifying} view this problem as a multi-armed bandit problem; they first cluster the samples in a pool, and select the samples from a group more often when more unknown unknowns are discovered from the group.
However, it is not clear whether this strategy yields better sampling efficiency in terms of the predictor's performance.

\citet{chen2018my} define unfairness as the difference in classification accuracy between two groups and suggest additional data collection as one of the remedies for such unfairness. However, they do not study an active sampling method to efficiently reduce the classification error or the unfairness.

In the extensions of our method, we maximize the prediction change (i.e., EDG\_ext1 in our experiments). The method is related to a type of active learning strategies based on maximizing model change~\citep{settles2008multiple,settles2008analysis,zhang2015active}. However, these methods do not address the practical challenges such as transparency, applicability to black-box models, and robustness to labeling noise.

\section{Method}

\label{sec:method}
The main goal of batch active learning is to reduce the error $E(C,D_U)$ of a classifier or tagger $C$ on an unseen testing dataset $D_U$ after labeling a fixed number of samples. To simplify the explanation, we first assume that testing error can be well-approximated by validation error, and the methods that do not require validation data will be described later as extensions in Section~\ref{sec:extensions}.


Based on external datasets, previous work maximizes the error reduction by modeling interactions among various types of signals such as uncertainty estimation, input features, and the state of the sampling process~\citep{KonyushkovaSF17,bachman2017learning,fang2017learning,ravi2018meta}.
In order to not rely on a large external dataset, we first cluster samples into multiple groups, 
and assume that the validation error of the samples in each group only depends on the number of annotated samples in the same group. Then, estimating the error reduction is decomposed into simple one-dimensional regression problems which can be done by observing only a few pairs of validation errors and its corresponding number of samples.

In the following subsections, we describe our framework in the context of solving NER problems. The framework is generally applicable to any classification or sequential tagging problems if the methods used to cluster samples (Section~\ref{sec:clustering_feature_all}) and the error decay function (Equation~\ref{eq:approx_error}) are modified properly.

\subsection{Error Partition}
Given a feature, we can derive a partition $p$ by clustering the samples into $J^p$ groups. 
For example, using the sentence embedding as our features, we can use K-means~\citep{macqueen1967some} to cluster every sentence in the corpus into multiple groups containing sentences with similar embeddings. Then, the testing error $E(C,D_U)$ can be partitioned using the sentence groups as:
\vspace{-1mm}
\begin{align}\label{eq:error}
E(C,D_U) = \sum\limits_{s_i \in D_U} \sum\limits_{j=1}^{J^p} P(g^p_j|s_i) \sum\limits_{l=1}^{|s_i|} \mathbbm{1}( y_{i,l} \neq \hat{y}^C_{i,l}), 
\end{align}
where $C$ is the current predictor, $s_i$ is $i$th sentence in the testing data $D_U$,  $P(g^p_j|s_i)$ is the probability that the sentence $s_i$ belongs to the $j$th group $g^p_j$, and $P(g^p_j|s_i)$ could be the indicator function $\mathbbm{1}(s_i \in g^p_j)$ if a hard clustering method is used. $\sum_{l=1}^{|s_i|} \mathbbm{1}( y_{i,l} \neq \hat{y}^C_{i,l})$ is the error of the sentence $s_i$ using predictor $C$, $|s_i|$ is the length of the sentence, $y_{i,l}$ is the ground-truth tag for the $l$th token in the sentence $s_i$, and $\hat{y}^C_{i,l}$ is the tag predicted by the predictor $C$. 



By assuming that the error of sentence $s_i$ could be approximated by the estimated average error of its groups $\hat{E}(g^p_j,C)$ (i.e., $\sum_{l=1}^{|s_i|} \mathbbm{1}( y_{i,l} \neq \hat{y}^C_{i,l})\approx \hat{E}(g^p_j,C) |s_i|$), we estimate the overall error as: 
\begin{align}\label{eq:feature_error}
\hat{E}^p(C,D_U)  = \sum\limits_{j=1}^{J^p} \hat{E}(g^p_j,C) m(g^p_j,D_U),  
\end{align}
where $m(g^p_j,D_U) = \sum_{s_i \in D_U} P(g^p_j|s_i) |s_i|$ could be viewed as the number of times group $g^p_j$ appears in $D_U$. 

In addition to sentence clusters, we can also rely on word features and word clusters to form a partition $p$. Consequently, the error becomes 
\begin{align}\label{eq:word_error}
E(C,D_U) = \sum\limits_{s_i \in D_U} \sum_{l=1}^{|s_i|} \sum\limits_{j=1}^{J^p} P(g^p_j|s_{i,l}) \mathbbm{1}( y_{i,l} \neq \hat{y}^C_{i,l}),  
\end{align}
where $s_{i,l}$ is $l$th token in the sentence $s_i$. Hence, $m(g^p_j,D_U) = \sum\limits_{s_i \in D_U} \sum\limits_{l=1}^{|s_i|} P(g^p_j|s_{i,l}) $ in Equation~\eqref{eq:feature_error}.







\subsection{Clustering for NER}
\label{sec:clustering_feature_all}



When we use different partitions $p$, we get different error estimates $\hat{E}^p(C,D_U)$. To increase the robustness of our error estimation, we adopt multiple partitions based on different features and aggregate the testing error estimates for selecting the next batch of samples to be annotated.

In our experiments on real-world NER datasets, 
we build four partitions using different features of sentences and words as follows:
\begin{itemize}[noitemsep]
\item Sentence: We compute sentence embeddings by averaging the word embeddings, and cluster all the sentence embeddings into 10 groups. Next, the cosine similarities between sentence embeddings and the cluster centers are passed through a softmax layer with temperature parameter $0.1$ to compute $P(g^p_j|s_i)$.
\item Word: We perform a simple top-down hierarchical clustering on word embeddings, which first clusters the words into 10 groups and further partitions each group into 10 clusters. This step results in 100 clusters for words in total.
\item Word + Shape: Instead of performing clustering on the lowest layer of the hierarchy, we partition the words in each group using four different word shapes: uppercase letters, lowercase letters, first uppercase letter and rest lowercase letters, and all the shapes other than above. The same word shape features are also used in our tagger.
\item Word + Sentence: Similarly, we partition each of the 10 word groups in the lowest layer of the hierarchy. For each word, we find the sentence $s_i$ the word belongs to, and rely on the sentence group $g^p_j$ with highest $P(g^p_j|s_i)$ to perform the partition.
\end{itemize}

Performing clustering on the concatenation of multiple feature spaces is less interpretable, so we choose to model the feature interdependency by hierarchical clustering (i.e., concatenating the clustering results). For example, in the third partition (i.e., Word + Shape), a cluster contains all words that have the same shape feature and belong to one of the 10 word embedding clusters.

Among the four partitions, the first one (i.e., Sentence) uses soft clustering because a sentence might contain multiple aspects that belong to different groups. We perform hard clustering on word features because it achieves similar performance when compared to soft clustering, and speeds up updating the cluster size when a new sample is added.
For efficiency, all the clustering is done by mini-batch K-means~\citep{sculley2010web} in $D_A$, the union of training data, sampling pool, and validation data. 

To simplify the method and to have a better control on experimental settings, we use the same method to form groups based on the tagger's input features for all datasets, and use the same groups when selecting all batches in a dataset. Nevertheless, we note that the framework allows us to model error decay on more fine-grained clusters as more training data are collected, or use other external features (e.g., the journal where the sentence is published) that might not be easily incorporated into the tagger or uncertainty sampling. 



\subsection{Error Decay Modeling}

Within each group, we assume that the error depends only on the number of samples in the group being observed in the training data. This is to avoid a complicated and uninterpretable error decay model built by many pairs of training data subset and validation error.
We model the error of predictor $C_{T_t}$ on $j$th group $\hat{E}(g^p_j,C_{T_t})$ in Equation~\eqref{eq:feature_error} using a one-dimensional function $e(n)$, where $n=m(g^p_j,T_t)$ is the size of group $g^p_j$ in the training data $T_t$ after $t$th batch is collected, and further constrain the class of decay functions $e(n)$ using prior knowledge of the tasks.



The decay function of prediction error $e(n)$ depends on the task~\citep{amari1992four,hestness2017deep}. The error decay rate of many tasks has been shown to be $1/n^k$, both theoretically and empirically~\citep{hestness2017deep}, and $k$ is typically between 0.5 and 2~\citep{amari1992four}. 

In sequence tagging tasks, the error decay rate depends on the importance of context. 
To intuitively explain how the importance of context affects the error decay rate, we discuss the form of error decay functions in one case where context does not affect the label and in another case where context matters. 

\noindent \textbf{Case 1 (Context does not Matter):}
Assuming we are classifying each token in a sentence into two classes and its label does not depend on context (like predicting the outcome of a coin toss), we only make reducible errors when we observe the less-likely label more times than the other label. Applying Chernoff bounds~\citep{mitzenmacher2017probability}, we can show that the error decay rate is as fast as an exponential function.

Without loss of generality, we assume the probability $q$ of observing positive class (i.e., head) in the $i$th token is smaller than 0.5. Let $n$ be the number of coin tosses be $n$ and the random variable  $X^i_j=\mathbbm{1}(\text{jth toss on ith coin is head} )$. In order to classify the testing tokens optimally (i.e., predict tail whenever seeing the $i$th token), we would like to observe $X^i = \sum_{j=1}^{n}X^i_j < \frac{n}{2} $. Therefore, the error rate is $P(X^i \geq \frac{n}{2})(1-q)+q(1-P(X^i \geq \frac{n}{2}))=P(X^i \geq \frac{n}{2})(1-2q)+q$.

Since all $X^i_j$ are assumed to be independent, we can use Chernoff bounds to model the decay of $P(X^i \geq \frac{n}{2})= P(X^i \geq (1+\delta)\mu )$ as $n$ increases, where $\mu=q\cdot n$, and $\delta=\frac{0.5-q}{q}$. Chernoff bounds tell us that $P(X^i \geq \frac{n}{2}) \leq \exp(-n \cdot h(q))$, where $h(q)$ is an error decay speed function that depends on $q$. Different versions of Chernoff bounds lead to different $h(\cdot)$, but all $h(\cdot)$ increase as $q$ decreases. That is, when coins are more biased, error rate decays faster.

\noindent \textbf{Case 2 (Context Matters):}
If we assume that the influence of each word in the context to the label is independent, we need to estimate the probability of having the label given a word in the context to predict the label accurately. For example, we want to know how likely the label is a person name when we observe "Dr" in the context, so that we can estimate how likely the Pepper in "Dr Pepper" should be labeled as a person. 
The error of the probability estimation decays with rate $1/\sqrt{n}$ in the long run according to Chernoff bounds or the central limit theorem, so the error decay function is likely to be as slow as $1/\sqrt{n}$ when the words in context affect the label.


The error decay rate of most of the NER tasks should lie between the decay rates in the above two cases because the taggers will gradually learn to utilize longer contexts. 
Thus, we model the error decay $\hat{E}(g^p_j,C_{T_t}) = e\left(m(g^p_j,T_t)\right)$ for NER by a fractional polynomial: 
\begin{align}\label{eq:approx_error}
 e(n) = c_j + b_j \left( \frac{a_{0.5}}{ (a_0 \cdot n)^{0.5}} + \sum\limits_{k=1}^3 \frac{a_k}{ (a_0 \cdot n)^k }  \right),
\end{align}
where $a_{0.5}$, $a_{0-3}$, $b_j$ and $c_j$ are parameters to be optimized, and we constrain these parameters to be non-negative. $c_j$ is an estimate of irreducible error in this model, $b_j$ tries to predict the initial error (when the first batch is collected), $a_{0.5}$ and $a_{1-3}$ weight the curves with different decay rates, and $a_0$ scales the number of training samples. If $e(n)$ is proportional to $1/\sqrt{n}$, the estimated $a_{1-3}$ would be close to 0. If $e(n)$ is proportional to an exponential function, $a_{1-3}$ would become the coefficients in its Taylor expansion.

The parameters $a_{0.5}$, $a_{0-3}$, $b_j$, and $c_j$ are estimated by solving
\begin{align}\label{eq:error_opt}
& \argmin_{\substack{ \{a_{0-3},a_{0.5}, \\
b_j,c_j\} \in \mathbf{R}^M_{\geq 0} }}  \sum_{j=1}^{J^p} w_j \sum_{t=1}^{t_m} v_{tj} \left( \hat{E}(g^p_j,C_{T_t}) - \frac{E^p_j(C_{T_t},D_V)}{m(g^p_j,D_V)} \right)^2, 
\end{align}
where $t_m$ is the number of annotated batches and the estimated error $\hat{E}(g^p_j,C_{T_t})=e\left( m(g^p_j,T_t) \right)$. The average error of $j$th group in validation dataset $D_V$, \\ $E^p_j(C_{T_t},D_V) = \sum_{s_i \in D_V} P(g^p_j|s_i) \sum_l \mathbbm{1}( y_{i,l} \neq \hat{y}^{C_{T_t}}_{i,l})$ for a partition using sentence clusters and $E^p_j(C_{T_t},D_V)= \sum_{s_i \in D_V} \sum_l P(g^p_j|s_{i,l})  \mathbbm{1}( y_{i,l} \neq \hat{y}^{C_{T_t}}_{i,l})$ for a partition using word clusters.
$w_j$ and $v_{tj}$ are constant weights\footnote{See supplementary material~\ref{sec:implement_details} for details on setting the weights}, and $M=2J^p+5$ is the number of parameters. Due to the small number of parameters, error decay curves could be modeled by retraining deep neural networks only a few times (we set $t_m=5$ when selecting the first batch in our experiments).





\subsection{Query Batch Selection}
Modeling the error decay on each cluster based on different features could be used as an analysis tool to increase the transparency of existing active learning methods. Such an analysis reveals the weaknesses (i.e., the groups of samples with high validation error) of the current tagger and allows us to estimate the number of samples that needs to be collected to reach a desirable error rate. 

We propose a novel active learning method to actively address the fixable weaknesses of the tagger discovered by the analysis tool. When a single partition $p$ is used, we select the next batch $B$ by maximizing
\begin{align}\label{eq:obj}
H^p(B \cup T) = - \sum\limits_j e\left(m(g^p_j,B \cup T)\right) m(g^p_j,D_A),
\end{align}
where $T$ is the collected training data, and $D_A$ is the union of the pools of candidate samples, training data $T$, and validation data $D_V$, which are used to approximate group occurrence statistics in the testing data $D_U$. Note that we use $e(m(g^p_j,B \cup T))$ to approximate $\hat{E}(g^p_j,C_{B \cup T})$, so we prevent retraining the predictor $C$ within each batch selection.





\textbf{Proposition 1.} 
\textit{Suppose that $\hat{E}(g^p_j,C_{T_t})$ is a twice differentiable, non-increasing and convex function with respect to $m(g^p_j,T_t)$ for all $j$, then $H^p(T)$ is non-decreasing and submodular.}

The convexity of $\hat{E}(g^p_j,C_{T_t})$ is a reasonable assumption because the error usually decays at a slower rate as more samples are collected. Since selecting more samples only decreases the value of adding other samples, $H^p(T)$ is submodular (see supplementary material~\ref{sec:prop1_proof} for a rigorous proof).

Finding the optimal $B$ in Equation~\eqref{eq:obj} is NP-complete because the set cover problem can be reduced to this optimization problem~\citep{guillory2010interactive}, but the submodularity implies that a greedy algorithm could achieve $1-1/e$ approximation, which is the best possible approximation for a polynomial time algorithm (up to a constant factor)~\citep{lund1994hardness,guillory2010interactive}.

\begin{algorithm*}[!t]
\SetAlgoLined
\SetKwInOut{Input}{Input}
\SetKwInOut{Output}{Output}

\Input{Sampling pool without labels, validation data with labels $D_V$, a predictor $C$, number of clusters $J^p$ for every partition $p$, burn-in batch number $t_{b}$, total batch number $t_{max}$}
\Output{Labels for selected batches}

\ForEach{partition $p$}{%
    Cluster samples into groups $g^p_j$, where $1 \leq j\leq J^p$ 
}

\For{$t_m\gets 1$ \KwTo $t_{max}$}{
    \uIf{$t_m \leq t_{b}$}{%
        Randomly sample batch $B$
    }
    \Else{%
    	Model error decay by solving~\eqref{eq:error_opt} \\
		Select batch $B$ by solving~\eqref{eq:selection_features}
    }
    Add the batch $B$ to training data $T_{t_m}$, remove the batch $B$ from sampling pool \\
	Train the predictor $C_{T_{t_m}}$ \\
	Test the predictor on validation data to compute $E^p_j (C_{T_{t_m}}, D_V), \forall j, p$
}
 \caption{Error decay on groups (EDG) selection algorithm}
 \label{algo:edg}
\end{algorithm*}

When having multiple partitions $p$ based on different features, we select the next sentence in the batch according to:
\vspace{-1mm}
\begin{equation}\label{eq:selection_features}
\argmax\limits_{s_i} \left( \prod\limits_{p} \left( \frac{H^p( S_i \cup T) - H^p(T)}{|s_i|} + \epsilon \right) \right)^{\frac{1}{F}},
\end{equation}
where $\epsilon$ is a small smoothness term. If the partition $p$ is a set of sentence clusters, then $S_i = \{s_{i}\}$. If it is is a set of word clusters, then $S_i = \{s_{i,l}\}_{l=1}^{|s_i|}$, where $s_{i,l}$ is $l$th word in $i$th sentence. 
We normalize the error reduction $H^p( S_i \cup T) - H^p(T)$ by the sentence length $|s_i|$ to avoid the bias of selecting longer sentences as done in previous work~\citep{settles2008analysis,shendeep}. After annotators label the whole batch, we retrain the tagger model and update the error decay prediction by solving~\eqref{eq:error_opt} before selecting the next batch. 

The selection process is summarized in Algorithm~\ref{algo:edg}. In the first few batches, we perform random sampling to collect pairs of every cluster size and the prediction error in the cluster for solving the one-dimensional regression problems. After the number of collected batches $t_m$ is larger than the number of burn-in epochs $t_{b}$, we have sufficient size and error pairs required to model the error decay in~\eqref{eq:error_opt}. Then, we can predict the future error $H^p( S_i \cup T)$ and select the samples that minimize the error using~\eqref{eq:selection_features}.


Note that~\eqref{eq:selection_features} naturally balances informativeness and representativeness. From the informativeness perspective, the sample without error decay won't be selected. From the representativeness and diversification perspectives, we will decrease the value of choosing a sample in a batch after the samples in the same clusters are selected.



\section{Method Extensions}
\label{sec:extensions}
For some applications, a validation set is not large enough to be used to model the error decay curves, and our independent assumption may be too strong. To address this concern, we also test two extensions of our method.

\subsection{Prediction Difference Decay} 
\label{sec:first_ext}
We replace the ground truth labels in~\eqref{eq:error_opt} with the prediction $\hat{y}^{C_{T_{t_m}}}_{i,l}$ based on the current training data. That is, our sampling method computes the difference between the current prediction and the previous predictions $C_{T_{1}}$, ..., $C_{T_{t_m-1}}$  in each group, and models the decay of the difference to maximize the convergence rate of predictions. We denote this method as EDG\_ext1 in our experiments.
\subsection{Uncertainty or Disagreement Decay}
\label{sec:second_ext}

When the uncertainty or disagreement information is available, we can model their decay and choose the sentences with highest uncertainty decay rather than highest uncertainty. To avoid making the independence assumption, we skip the clustering step and assume that future uncertainty decay is proportional to the previous uncertainty decay, and set the score of $i$th sentence to be 
\vspace{-1mm}
\begin{align}\label{eq:UD}
\min( \max(u^{t_f}_i-u^{t_m}_i,0), u^{t_m}_i),
\end{align}
where $u^{t_m}_i$ is the current uncertainty of $i$th sentence, and $u^{t_f}_i$ is its previous uncertainty. Note that we take the minimum between the difference and $u^{t_m}_i$ to ensure that the predicted future uncertainty is always non-negative. This method is denoted as EDG\_ext2 in our experiments.


\section{Experimental Setup}
NER problems are often used as benchmark to evaluate (deep) active learning methods~\citep{settles2008analysis,shendeep,siddhant2018deep} because they are the foundation for many information extraction tasks, and acquiring tags for each token requires a large amount of human effort. 
We follow \citet{StrubellVBM17} and use phrase-level micro-F1 as the performance metric for NER tasks. 
Precision and recall are computed by counting the number of correct boundary and type predictions.
Unless otherwise stated, we use a four-layer convolutional neural network (CNN) as our tagger~\citep{StrubellVBM17}\footnote{We choose to test the methods on CNN because \citet{StrubellVBM17} showed that CNN can achieve performance which is close to state of the art while being much more efficient than BiLSTM-CRF.}.

\subsection{Simulation on Gold Labels}
This is one of the most widely used setups to evaluate active learning methods. We compare the performance of NER tagger trained on different data subsets chosen by different methods. In the supplementary material~\ref{sec:more_exp_results}, we also compare the performance of applying different active learning methods to BiLSTM-CRF models.

\subsubsection{Synthetic Dataset}
We synthesize a dataset with 100 words; each word could be tagged as one of four entity types or none (not an entity). There are three categories of words. The first category consists of half of the words which are always tagged as none. This setup reflects the fact that a substantial amount of words such as verbs are almost always tagged as none in NER tasks. 

One-fourth of the words belong to the second category where every word mention has equal probability of being tagged as one of the entity types or none. In real-word NER tasks, the noisy label assignment may be due to inherently ambiguous or difficult words. 

The remaining 25 words are in the third category where the labels are predictable and depend on the other context words. The likelihood of words in the third category being tagged as one of the four entity types is sampled from a Dirichlet distribution with $\alpha_{1-4} = 1$, while the likelihood of being none is zero. Whenever one of these words $w$ appear in the sentence, we check two of its preceding and succeeding words that are also in the third category, average their likelihoods of entity types, and assign the type with the highest likelihood to the word $w$.

When generating a sentence, the first word is picked randomly. The transition probability within each category is $0.9$. Inside the first and second categories, the transition probability is uniformly distributed, while the probability of transition to each $w$ inside the third category is proportional to a predetermined random number between $0.1$ and $1$.
The sentence length is between $5$ and $50$ and there is a  $0.1$ probability of ending the sentence after generating a word within the range.

In this dataset, each word is a group in our method and no clustering is performed. The $i$th word has a word embedding vector $[\mathbbm{1}(k=i)]_{k=1}^{100}$. When modeling error decay, we start from 1,000 tokens and use a batch size of 500. When evaluating the sampling methods, we start with 3,000 using a batch size of 1,000. That is, after the first batch is selected, we update the error decay curves based on the prediction of taggers trained on 1,000, 1,500, 2,000, 2,500, 3,000 and 4,000 tokens. 



\begin{table}[t!]
\centering
\begin{tabular}{|c|cc|cc|}
 \hline 
    &  \multicolumn{2}{c|}{Synthetic data} & \multicolumn{2}{c|}{CoNLL 2003} \\
      & Token            & Sentence  & Token & Sentence          \\ \hline
Train &  99,956      & 6,726 &  204,567        & 14,987    \\
Val   & 10,045       & 679 &  51,578         & 3,466     \\
Test  & 10,004       & 677  &   46,666         & 3,684 \\ \hline     &  \multicolumn{2}{c|}{NCBI disease} & \multicolumn{2}{c|}{MedMentions ST19} \\
      & Token            & Sentence  & Token & Sentence          \\ \hline
Train &     135,900   &  5,725 &   758,449      & 28,227    \\
Val   &     23,836   & 941 & 254,539 &   9,303   \\
Test  &     24,255   &  970 & 253,737 & 9,383 \\ \hline
\end{tabular}
\caption{The size of datasets for the simulation on gold labels.}
\label{tb:dataset_size}
\end{table}

\subsubsection{Real-world Datasets}
We test the active sampling methods on CoNLL 2003 English NER~\citep{tjong2003introduction}, NCBI disease~\citep{dougan2014ncbi}, and MedMentions~\citep{Murty18} datasets. The size of these datasets are presented in Table~\ref{tb:dataset_size}. CoNLL 2003 dataset has four entity types: people name (PER), organization name (ORG), location name (LOC), and other entities (MISC). NCBI disease dataset has only one type (disease name). For MedMentions, we only consider semantic types that are at level 3 or 4 (higher means more specific) in UMLS~\citep{bodenreider2004unified}. Any concept mapping to more abstract semantic types is removed as was done by \citet{greenberg2018marginal} and this subset is called MedMentions ST19. The 19 concept types in MedMentions ST19 are \emph{virus, bacterium, anatomical structure, body substance, injury or poisoning, biologic function, health care activity, research activity, medical device, spatial concept, biomedical occupation or discipline, organization, professional or occupational group, population group, chemical, food, intellectual product, clinical attribute,} and \emph{Eukaryote}.

In all the three datasets, the first 30,000 tokens are from randomly sampled sentences. To model error decay, we start from 10,000 tokens and retrain the tagger whenever 5,000 new tokens are added. When evaluating the sampling methods, we start from 30,000 tokens, using a batch size of 10,000.\footnote{Selecting only part of a sentence is not reasonable in NER, so each chosen batch may be slightly larger than the desired batch size. The difference is smaller than the length of the last chosen sentence. Since the desired batch size is set to be large (10,000 words in real-world datasets), the difference in batch size between sampling methods is negligible.}

\begin{figure}[t!]
\centering
\includegraphics[width=0.7\linewidth]
{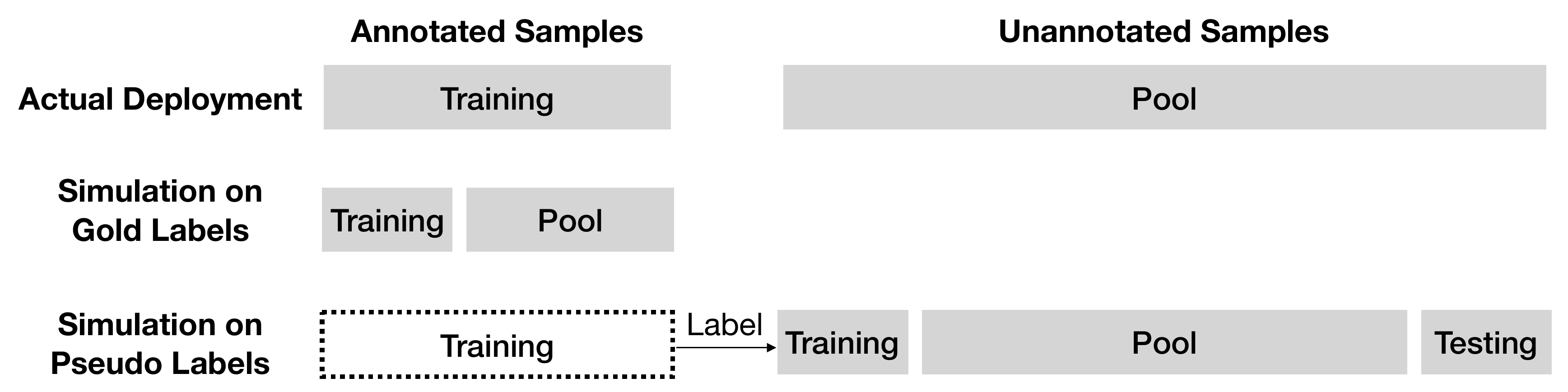}
\caption{Simulation on pseudo labels compares active learning methods on a large pool with noisy labels. In addition to the original validation and testing set, we also use a testing set with pseudo labels for evaluation.}
\label{fig:pseudo_experiment}
\end{figure}

\subsection{Simulation on Pseudo Labels}
In practice, we often observe systematic noise from annotators. The noise could come from some inherently difficult or ambiguous cases in the task or 
from incapable workers in the crowdsourcing platforms. Thus, we propose a novel evaluation method to test the robustness of different sampling methods in the presence of such noise. 

As shown in Figure~\ref{fig:pseudo_experiment}, we first train a high-quality tagger using all the training data and use it to tag a large sampling pool. Then, different active learning methods are used to optimize the tagger trained on these pseudo labels. The micro-F1 is measured by comparing the tagger's prediction with pseudo labels or gold labels on unseen sentences. The evaluation method also allows us to perform sampling on a much larger sampling pool\footnote{Running sampling methods on a large pool is time-consuming, so we only compare the methods after the first batch is collected, i.e., when the size of the training dataset reaches 40,000.}, which is usually used in the actual deployment of active learning methods. We randomly select 100,000 abstracts from PubMed as our new sampling pool, which is around 180 times larger than the pool we used in the simulation on gold labels. Precisely, the sampling pool consists of 24,572,575 words and 921,327 sentences. The testing data with pseudo labels have 2,447,607 tokens and 91,591 sentences.

We also evaluate the sampling methods on two practical variations of the above setting. We use the data collected for optimizing a CNN to train BiLSTM-CRF models, which can be used to test the robustness of active learning methods after switching tagger models~\citep{lowell2018transferable}. In addition, when collecting gold labels for biomedical NER, annotators often tag the whole abstract at a time, which can only be tested using a large sampling pool. 

For all sampling methods, the sampling score of an abstract is the average of the sampling scores of the sentences in the abstract weighted by the sentence length. That is, the selection criteria view an abstract as a bag of sentences similar to how a sentence is considered as a bag of words when clustering is performed on words. We greedily select the abstract with the highest sampling score.




\subsection{Sampling Strategies}









We compare the following sampling methods:

\begin{itemize}[noitemsep]
\item Random (\textbf{RND}): We select sentences randomly with uniform probability.
\item Error Decay on Groups (\textbf{EDG}): This is our method where we optimize~\eqref{eq:selection_features} using validation data.
\item \textbf{EDG\_ext1 (w/o Val)}: As described in Section~\ref{sec:first_ext}, we replace the validation error in EDG with the prediction difference.
\item Maximum Normalized Log-Probability (\textbf{US}): We use the least confidence sampling~\citep{culotta2005reducing}. This variant of uncertainty sampling has been shown to be very effective in NER tasks~\citep{shendeep}. When applying maximum normalized log-probability to the CNN model, we select the sentences via $\argmin\limits_{s_i} (1/|s_i|) \sum_{l=1}^{|s_i|} \max\limits_{y_{il}} \log P(y_{il})$.
\item Maximum Normalized Log-Probability with Diversification (\textbf{US + Div}): We diversify uncertain samples based on sentence embeddings (i.e., the average embedding of its words)~\citep{wei2015submodularity,shendeep}. We implement the US + Div, also called filtered active submodular selection (FASS), described in \citet{shendeep}. We use cosine similarity to measure the similarity between sentence embeddings. The number of candidate sentences is the batch size times $t=100$. 
\item Diversification (\textbf{Div}): We use the same algorithm as US + Div, except that all samples are equally uncertain.
\item \textbf{US + Div + EDG\_ext2}: This is the same algorithm as US + Div, but with uncertainty scores replaced with their difference in Equation~\eqref{eq:UD}.
\item Bayesian Active Learning by Disagreement (\textbf{BALD}): We select samples based on the disagreement among forward passes with different dropouts~\citep{gal2017deep}. The prediction disagreement of $l$th token in $i$th sentence is computed by 
$\frac{\sum_{k=1}^K \mathbbm{1}(y^k_{il} \neq \text{mode}_{k'}(y^{k'}_{il}) ) }{K}$. The number of forward passes $K$ is set to 10 in our experiments. The sentence disagreement is the average of tokens' disagreement.
We use the default hyperparameter values for the dropouts as in \cite{StrubellVBM17}.
\item \textbf{BALD + EDG\_ext2}: Here, disagreement scores are replaced with their difference in Equation~\eqref{eq:UD}.


\end{itemize}

\subsection{NER Tagger Details}
We use the published hyperparameters\footnote{\url{https://github.com/iesl/dilated-cnn-ner}} for real-world datasets, and simplify the tagger for the synthetic dataset to decrease the standard deviation of micro-F1 scores. In the synthetic dataset, we reduce the number of layers of CNN to two because the label depends only on the left two words and right two words. Furthermore, we change the learning rate from $5 \times 10^{-4}$ to $10^{-4}$, batch size from 128 to 32, and max epochs from 250 to 1000 to make the performance more stable. When training BiLSTM-CRF, we also use the implementation and its default hyperparameters from \citet{StrubellVBM17}. In all the experiments, the number of epochs is chosen using validation data.

The word embeddings for CoNLL 2003 are vectors with 50 dimensions from SENNA~\citep{collobert2011natural}. The word embeddings for NCBI disease and MedMentions ST19 are word2vec~\citep{mikolov2013distributed} with 50 dimensions trained on randomly sampled 10\% of all PubMed text. Before clustering, we normalize all the word embedding vectors such that the square of the $\ell_2$ distance between two words is twice their cosine distance. 
 
\subsection{Visualization of the Error Decay Model}

In addition to qualitatively evaluating our methods, we also visualize our error decay models. The visualization examines whether our error decay function in Equation~\eqref{eq:approx_error} can accurately model the empirical error decay in NER datasets, and whether our sample selection strategies are transparent and interpretable.

Given a dataset and a partition $p$, the empirical and predicted errors in each group $g^p_j$ are plotted as two curves in Figure~\ref{fig:error_curves}. We compare the y value of empirical error $E^p_j(C_{T_t},D_V)/m(g^p_j,D_V)$ and predicted error $\hat{E}(g^p_j,C_{T_t})=e\left( m(g^p_j,T_t) \right)$ given different x values $m(g^p_j,T_t)$, the cluster size of group $g^p_j$ in the training set. Each curve connects six points corresponding to $t=1...6$, and the total number of words in training set $|T_t|$ are 10,000, 15,000, 20,000, 25,000, 30,000, and 40,000 in real world datasets, respectively. The first 30,000 words are selected randomly and the last 10,000 words are selected by EDG.

In different datasets, we visualize the different partitions derived from different features. In synthetic data, a group is a word. In CoNLL 2003, we plot each group that contains all the words with the same shape. In NCBI disease, we plot each group that contains words with similar word embeddings. In MedMentions ST19, we plot each group that contains sentences with similar sentence embeddings. Note that in the quantitative experiments, we actually use word + shape in CoNLL 2003 and 100 clusters in NCBI disease, but we illustrate only shape in CoNLL 2003 and 10 clusters in NCBI disease to simplify the figure.

\section{Results and Analysis}

\begin{figure*}[t!]
\centering
\includegraphics[width=0.9\linewidth]
{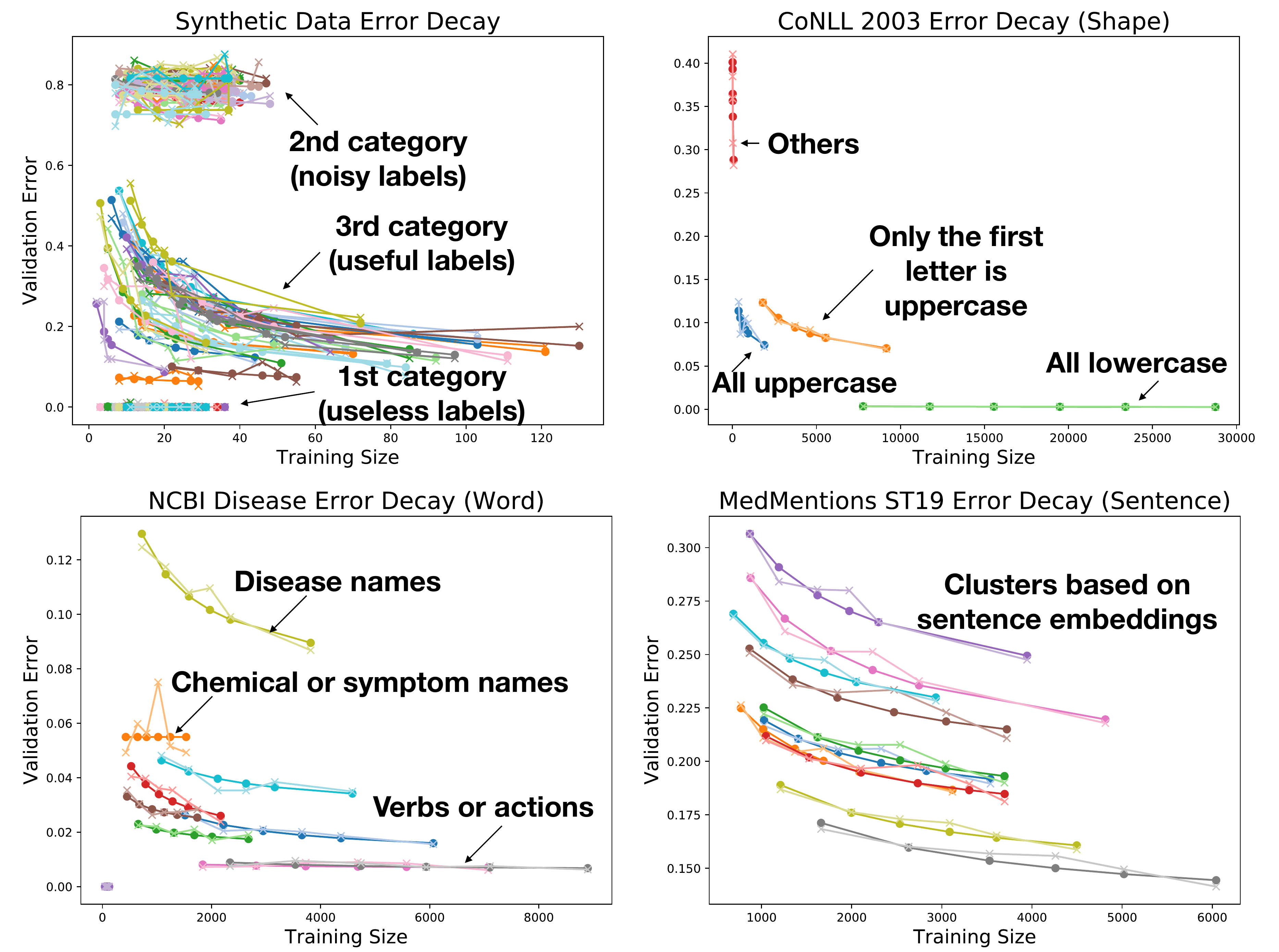}
\caption{Error decay on groups that are modeled after the first batch is collected by EDG. The \textbf{x} markers on the curves are the real error and $\bullet$ means prediction from the fitting curve. The groups shown in the figure for NCBI disease and MedMentions ST19 are formed by clustering word and sentence embeddings, respectively.}
\label{fig:error_curves}
\end{figure*}

\begin{figure*}[t!]
\includegraphics[width=\linewidth]
{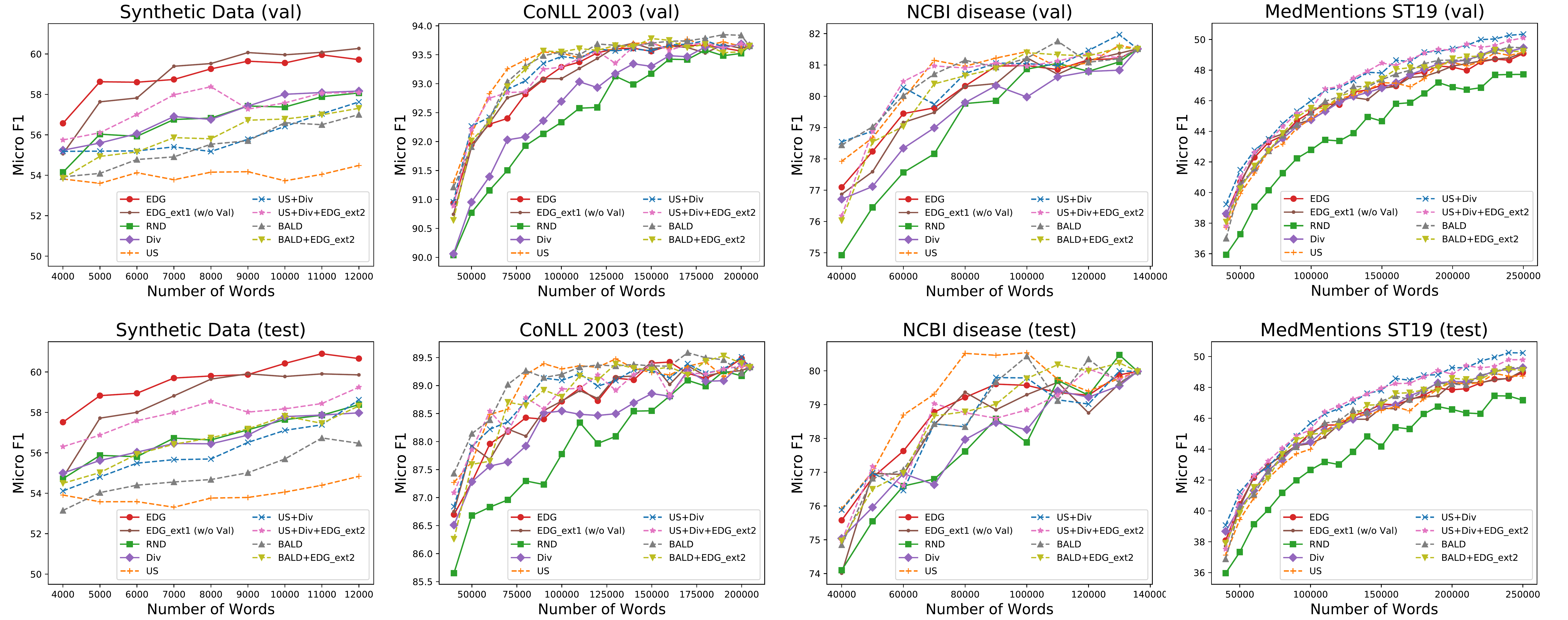}
\caption{Comparison of different sampling methods on the four NER tasks. The validation (first row) and testing scores (second row) are averaged from the micro-F1 (\%) of three CNNs trained with different random initializations. The performance of methods which cannot be applied to black-box taggers is plotted using dotted curves.}
\label{fig:gold_simulation}
\end{figure*}

The results of error decay visualization, the simulation on gold labels, and pseudo labels are shown in Figure~\ref{fig:error_curves}, Figure~\ref{fig:gold_simulation} and Table~\ref{tb:pseudo_exp}, respectively.

As shown in Table~\ref{tb:pseudo_exp}, the scores on the gold validation set, the gold testing set, and the testing set using pseudo labels follow a similar trend and most active learning methods do better than random regardless of which test set is used. The observation indicates that the taggers do not overfit the label noise in the training data severely and justifies our pseudo label experiments.

We first qualitatively analyze the error decay modeling in Section~\ref{sec:error_decay_vis}. Next, we quantitatively compare different methods in Sections~\ref{sec:edg_vs_div},
\ref{sec:us_vs_usdiv}, \ref{sec:edg_vs_us}, and \ref{sec:edg_ext1_vs_edg}.

\subsection{Error Decay Visualization}
\label{sec:error_decay_vis}
In Figure~\ref{fig:error_curves}, the predicted error decay curves usually fit the empirical values well. Some deviations of empirical error come from the randomness of training CNNs. For example, the fourth point in MedMentions ST19 has higher empirical error in almost all clusters because the parameters of CNN trained on the set with 25,000 words happen to converge to a worse state. 

In the figures, we can see that the length between the fifth and sixth points in each curve varies because the last 10,000 words in the training set are actively selected. The clusters that might have a larger error decay (e.g., the orange curve in CoNLL 2003) would get more training instances in the last sample batch. The figures demonstrate that the one-dimensional regression problem for each cluster could be solved well even though the sampling process is not random and only six training pairs for each curve are observed.




%

 We can interpret the sampling strategy of EDG from the different number of samples selected in different groups. For example, EDG improves the sampling efficiency when compared to random sampling by selecting more words whose first letter is uppercase in CoNLL 2003, and by selecting more disease name candidates than verbs or actions in NCBI disease.

The transparency of EDG explains some important empirical observations in previous work. For example, \citet{lowell2018transferable} observed that the benefit of active learning is usually more significant in NER than in sentence classification, and the improvement in NER is robust against the change of predictor model.
\citet{shendeep} observed that in some NER datasets (e.g., CoNLL 2003), we can train a neural tagger that reaches a similar performance using only a selected small portion of the training set compared to using all the data.  Figure~\ref{fig:error_curves} indicates that one of the main reasons is that there are more lowercase words in CoNLL 2003 than uppercase words, and lowercase words are almost never tagged as names of people, organizations, or locations. Therefore, the active learning methods could easily achieve high sampling efficiency by selecting more uppercase words, and the selection tendency can benefit various kinds of predictor models.




In real-world datasets, the error decay rate usually follows the function $1/\sqrt{n}$ when $n$ is large for most of the groups regardless of the feature being used.  For example, $a_{0.5}$ in Figure~\ref{fig:error_curves} is at least two times larger than the $a_{1}+a_{2}+a_{3}$ in CoNLL 2003, NCBI disease, and MedMentions ST19. 
The small difference between empirical and predicted error also justifies our assumption that the weight parameters of terms are shared across all the groups (i.e., $a_{0-3}$ and $a_{0.5}$ do not depend on the cluster index $j$).

\begin{table*}[t!]
\centering
\scalebox{0.65}{
\begin{tabular}{|c|ccc|ccc|ccc|ccc|c|}
\hline
& \multicolumn{6}{c|}{Whole abstract} & \multicolumn{6}{c|}{Sentence} & \multirow{4}{*}{Avg} \\ \cline{2-13}
 & \multicolumn{3}{c|}{\multirow{2}{*}{CNN for CNN}} & \multicolumn{3}{c|}{CNN for} & \multicolumn{3}{c|}{\multirow{2}{*}{CNN for CNN}} & \multicolumn{3}{c|}{CNN for}&  \\
& \multicolumn{3}{c|}{} & \multicolumn{3}{c|}{BiLSTM-CRF} & \multicolumn{3}{c|}{} & \multicolumn{3}{c|}{BiLSTM-CRF} &                 \\  \cline{2-13}
& Val& Test& Pseudo&Val& Test& Pseudo&Val& Test& Pseudo&Val& Test& Pseudo &         \\ \hline
EDG$\dagger$& \textbf{61.0}& \textbf{59.2} &\textbf{54.9} &   \textbf{66.5}&\textbf{66.8}&\textbf{58.7} &
60.1 &58.2 & \textbf{54.7} & \textbf{68.2}&66.5&\textbf{59.1}&\textbf{61.2}\\
EDG\_ext1 (w/o Val) &59.6&58.5&\textbf{54.9}&65.7&65.5&57.7&\textbf{60.2}&\textbf{58.4}&52.5&67.7&\textbf{67.8}&58.5&  60.6 \\
RND&56.0&56.3&50.4&62.6&62.6&55.0&57.7&56.1&52.9&63.9& 64.1&56.2& 57.8 \\
Div & 55.2 & 55.6& 52.8 & 63.2 &63.9&56.1& 57.7 & 56.2 & 52.5 &63.9 & 64.2 & 56.6  & 58.2          \\ \hline
US&59.7& 59.0 &54.1 &    
67.2 & 67.4 & 57.8 & 58.3&57.7&53.1 &68.5 &\textbf{68.0} &58.1 &60.7
\\
US+Div&  60.1 &\textbf{59.5} & \textbf{56.0}  & 64.1&65.7&57.9 & 60.1 & 56.8 &54.5 & 68.7 &67.4& 58.5&60.8 \\
US+Div+EDG\_ext2 & \textbf{61.2} & 58.6 &	\textbf{56.0}&\textbf{69.5} & 65.4 & 58.7 &62.2 & 59.4 & \textbf{55.9} & 69.8 & 66.6 & 58.5 & 61.8 \\
BALD & 59.8 & 59.4 & 55.5 & 67.4 & \textbf{67.7} &	58.3&
59.9 & 58.8 & 54.9 & 
69.7 & 67.3 & \textbf{59.5} & 61.5 \\
BALD+EDG\_ext2 & 60.7 & 58.2 & \textbf{56.0} &
68.4 &65.8 & \textbf{58.9}&
\textbf{63.7} & \textbf{60.0} & 54.9 & 
\textbf{71.7} & 67.6 & 59.2 &	\textbf{62.1}
\\
 \hline
\end{tabular}
}
\caption{Simulation on pseudo labels for NCBI disease dataset. After selecting the first batch using different sampling methods, the micro-F1 (\%) is computed by averaging across five neural networks trained using different random initializations. Whole abstract and sentence mean we sample one abstract and sentence at a time, respectively. CNN for BiLSTM-CRF means that we report the F1 of BiLSTM-CRF that is trained on the data selected for optimizing CNN. The highest score for black-box and uncertainty-based models are highlighted, and the last column shows the unweighted average of all values in each row. The average F1 difference of EDG\_ext1 vs Div, US+Div+EDG\_ext2 vs US+Div, BALD+EDG\_ext2 vs BALD are significant ($p<0.01$) according to two-sample t-test. $\dagger$ indicates the method uses ground-truth labels in the validation set to collect samples in the training set.}
\label{tb:pseudo_exp}
\end{table*}



\subsection{EDG vs Div and RND}
\label{sec:edg_vs_div}
Among the methods we compared against, only random (RND) and diversification (Div) can be applied to black-box taggers. Our method (EDG) significantly outperforms Div, and Div outperforms RND in synthetic, CoNLL 2003, and NCBI disease datasets, which demonstrates the effectiveness of EDG. This also justifies our assumptions and indicates that the error decay curves are modeled well enough for the purpose of active sampling.


\subsection{US vs US+Div}
\label{sec:us_vs_usdiv}

\citet{shendeep} found that diversification is surprisingly not helpful in batch active learning. However, our results suggest that this finding might be valid only when the sampling pool size is small and/or some groups of frequent words/sentences are clearly not helpful.
When the pool is sufficiently large and the task is to jointly extract many different types of entities like MedMentions ST19, sampling almost all kinds of sentences can be helpful to the task as all sentence clusters have similar decay on the far right of Figure~\ref{fig:error_curves}. Then, the diversification approach (Div) can be as effective as US and EDG, while US+Div(+EDG\_ext2) provides the best result.

\subsection{EDG vs Uncertainty-based Methods}
\label{sec:edg_vs_us}

As shown in \citet{wang2017active}, it is difficult to perform better in terms of sampling efficiency when comparing a black-box active learning method with uncertainty sampling (US). In real-world datasets, EDG achieves part of the performance gain from US that is easily explainable (e.g., coming from ignoring those easy words), controllable by humans\footnote{In the supplementary material~\ref{sec:weighted_class}, we show that EDG can be easily modified for situations where each label class has a different penalty.}, and does not involve the specifics of the tagger to model the interaction between each word and its context.


Furthermore, US is not robust to labeling noise or ambiguous samples~\citep{mussmann2018relationship}, which have high errors but low error decay. For instance, US almost always selects difficult words with high irreducible errors in the synthetic data. In real-world dataset, we could also observe ambiguous or difficult words. For example, \emph{insulin} is a chemical, but \emph{insulin resistance} could be a disease or a symptom in NCBI disease dataset. In Figure~\ref{fig:error_curves}, we can see that EDG does not select many words in the group. Our error curves in the supplementary material~\ref{sec:US_analysis} show that US selects many such words with incorrect pseudo labels. The vulnerability makes EDG outperform US in synthetic data and Table~\ref{tb:pseudo_exp} on average.
 



US+Div and BALD are more robust to labeling noise than US, but still suffers from a similar problem. Thus, the sampling strategies that choose more samples with reducible uncertainty (i.e., US+Div+EDG\_ext2 and BALD+EDG\_ext2) could significantly improve the accuracy of taggers in noisy datasets like our synthetic data and NCBI disease dataset with pseudo labels, while having comparable performance on the other clean datasets with gold labels.






\subsection{EDG\_ext1 (w/o Val) vs EDG}
\label{sec:edg_ext1_vs_edg}
In all datasets, modeling the error decay using pseudo labels (EDG\_ext1)  achieves similar performance when compared to using gold validation data (EDG), and also outperforms Div. In addition, the micro-F1 scores of EDG on validation and testing data roughly show a similar trend, which suggests that our method does not overfit the validation data even though it has access to its gold labels during sampling.

\section{Conclusions}
We proposed a general active learning framework which is based only on the predictions from black-box predictors, is robust to labeling noise without relying on prior knowledge, and forecasts the potential error reduction in different aspects based on a small number of existing labels. 

Our experimental results suggest that no single batch active learning method wins in all the cases and every method has its own weaknesses. We recommend practitioners to analyze the error decay on groups in order to choose a proper sampling algorithm. If the sampling pool is small and the error decay analysis shows that many samples could be easily tagged, then uncertainty sampling methods are expected to perform well. Otherwise, diversification should be considered or combined with uncertainty sampling. Finally, error decay on groups (EDG) or its extensions should be adopted if there are practical deployment challenges such as issues of \emph{applicability} (e.g., only a black-box predictor is available), \emph{robustness} (e.g., labels are inherently noisy), or \emph{transparency} (e.g., an interpretable sampling process or an error reduction estimation is desired).

\section{Future Work}
\label{sec:future_work}
In our experiments, we demonstrated that our methods are transparent and robust to labeling noise. However, we have not yet applied them to tasks other than NER. 
For example, when we annotate a corpus for relation extraction, we usually want to select a document which is informative for the named entity recognizer, entity linker, and sentence classifier. This challenge is also called multi-task active learning~\citep{reichart2008multi,settles2011theories}. Compared to heuristically combining uncertainty from different models~\citep{reichart2008multi}, our methods provide more flexibility because it allows us to assign weights on the error reduction of each task and select the next batch by considering all tasks jointly.

In addition to the above pipeline system, question answering (QA) is another example where uncertainty is difficult to estimate. Many reading comprehension models such as pointer networks predict the start and end positions of the answer in a paragraph~\citep{wang2017gated}. However, higher uncertainty on the position prediction does not necessarily mean the model is uncertain about the answer. It is possible that the correct answer appears in many places in the paragraph and the network points to all the right places with similar low probability. By modeling the error decay directly, our methods avoid the issue.

Finally, we have not compared EDG with active learning methods that are designed for a specific task to solve a specific practical issue. For example, the active sampling methods proposed by \citet{wang2017active} are designed for semantic role labeling and focus on the applicability issue (i.e., black-box setting). Due to the difficulty of adapting their methods to NER and making fair comparisons, we leave such comparisons for future work.


\begin{acknowledgements}
We thank Akshay Krishnamurthy for many helpful discussions. We also thank the anonymous reviewers for their constructive feedback.

This work was supported in part by the Center for Data Science and the Center for Intelligent Information Retrieval, in part by the Chan Zuckerberg Initiative under the project “Scientific Knowledge Base Construction, in part using high performance computing equipment obtained under a grant from the Collaborative R\&D Fund managed by the Massachusetts Technology Collaborative, in part by the National Science Foundation (NSF) grant numbers DMR-1534431 and IIS-1514053. 

Any opinions, findings and conclusions or recommendations expressed in this material are those of the authors and do not necessarily reflect those of the sponsor.

This is a pre-print of an article published in Springer Machine Learning journal. The final authenticated version is available online at: \url{https://doi.org/10.1007/s10994-020-05897-1}

\end{acknowledgements}

%
%

\bibliographystyle{spbasic}      
\bibliography{ref}   

\newpage

\appendix

\section{Additional Experimental Results}
\label{sec:more_exp_results}

\subsection{CNN for BiLSTM-CRF using Gold Labels}

\citet{baldridge2004active} and \citet{lowell2018transferable} demonstrate that samples collected to optimize one model might not be helpful to another model. To test the robustness of different active learning methods to such model switch, we train BiLSTM-CRF models using the batches collected based on CNN (i.e., CNN for BiLSTM-CRF in Table~\ref{tb:pseudo_exp}). When training BiLSTM-CRF on the synthetic dataset, we increase the max epochs from 250 to 1000.

\begin{figure*}[t!]
\includegraphics[width=1\linewidth]{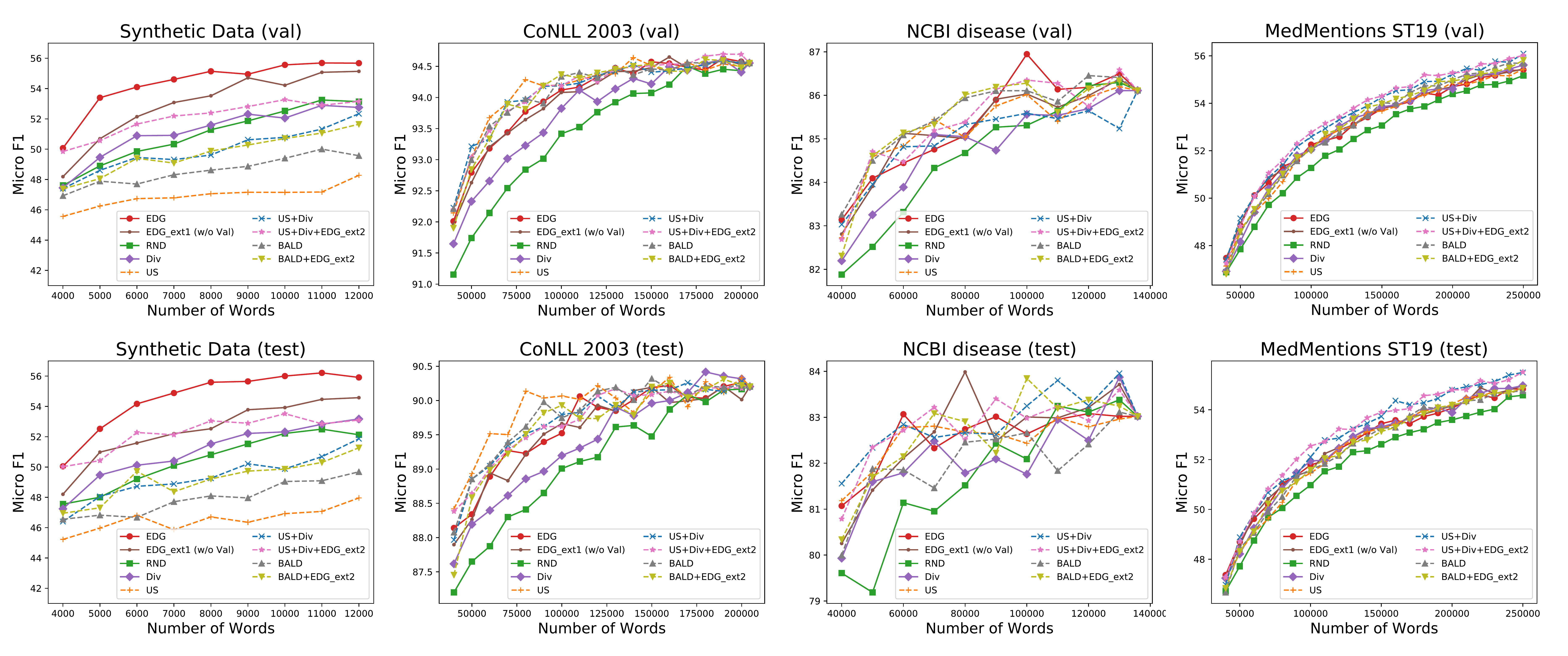}
\caption{Performance of BiLSTM-CRF models trained on training sets in Figure~\ref{fig:gold_simulation} (i.e., CNN for BiLSTM-CRF setting). The performance metrics are the average micro-F1 (\%) of three BiLSTM-CRF trained with different random initializations.}
\label{fig:CNN_for_BiLSTM-CRF}
\end{figure*}

The results are presented in Figure~\ref{fig:CNN_for_BiLSTM-CRF}. Nearly all the observations from Figure~\ref{fig:gold_simulation} also hold in Figure~\ref{fig:CNN_for_BiLSTM-CRF}. One difference is that EDG and its extensions perform slightly better. For example, in Figure~\ref{fig:gold_simulation} (CNN for CNN setting), BALD+EDG\_ext2 sometimes performs slightly worse than BALD (e.g., in the testing data of CoNLL 2003 and NCBI disease). However, BALD+EDG\_ext2 seems to always perform similar or better (e.g., in MedMentions ST19 and testing data of NCBI disease) than BALD after we switch the tagger model.

Another difference is that the performance gain between active sampling and random sampling is smaller in MedMentions ST19 than the gap in Figure~\ref{fig:gold_simulation}, while we do not observe a similar reduction in CoNLL 2003 and NCBI disease. We hypothesize that active learning methods will skip the groups of unhelpful words in CoNLL 2003 (like lowercase words) and NCBI disease, and those words are usually also unhelpful to other models. Thus, the performance gains in such datasets are more transferable and less dependant on the model choice in the first place.


\subsection{BiLSTM-CRF for BiLSTM-CRF using Gold Labels}

\begin{figure*}[t!]
\includegraphics[width=1\linewidth]{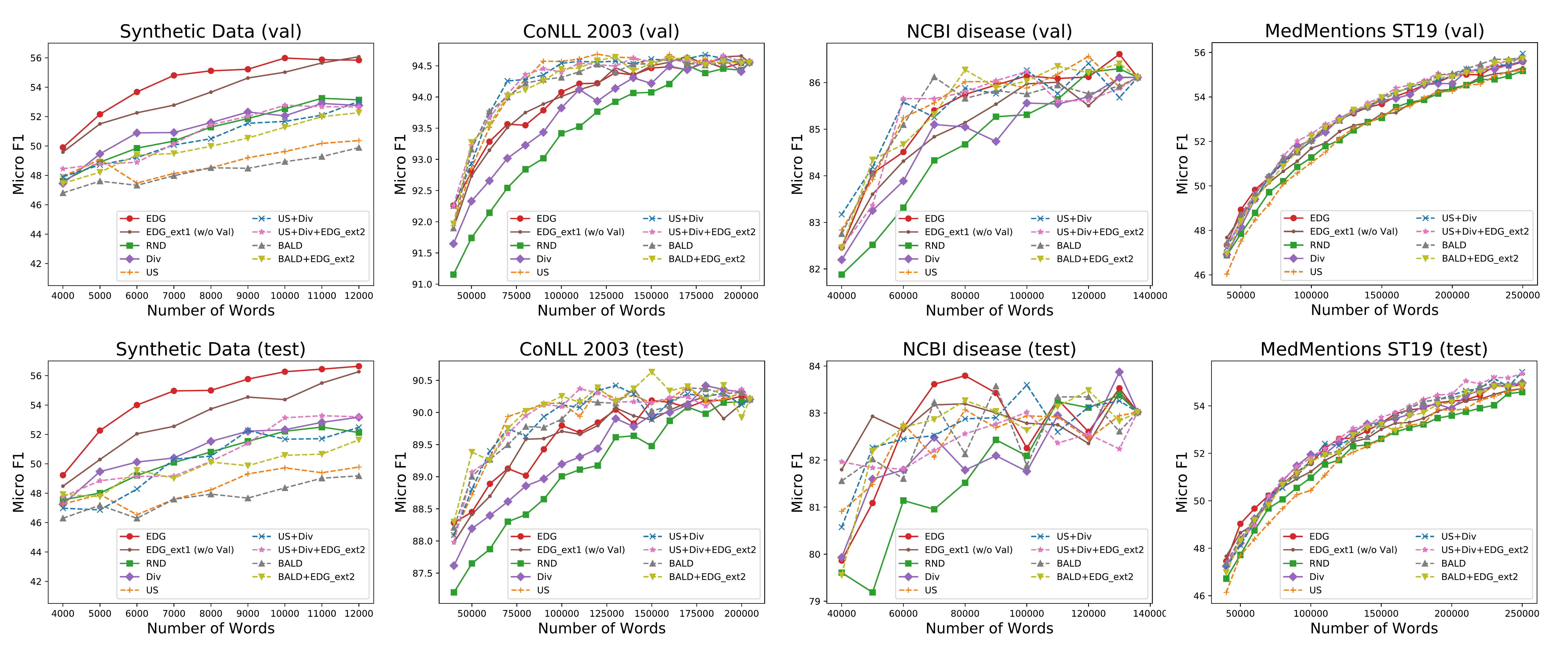}
\caption{Comparison of applying different sampling methods to BiLSTM-CRF models. The performance metrics are the average micro-F1 (\%) of three BiLSTM-CRF trained with different random initializations.}
\label{fig:CNN_for_BiLSTM-CRF_collect}
\end{figure*}

Following \citet{shendeep}, we set the uncertainty of each sentence as the negative log likelihood of the predicted label sequence. The results are presented in Figure~\ref{fig:CNN_for_BiLSTM-CRF_collect}. Most of the observations from Figure~\ref{fig:gold_simulation} also hold in Figure~\ref{fig:CNN_for_BiLSTM-CRF_collect}. For example, EDG and its extensions significantly improve the performances in our synthetic dataset, and are significantly better than diversification methods in CoNLL 2003 and NCBI disease datasets. There are some minor differences. For instance, the performance gap between EDG and uncertainty-based methods is larger in CoNLL 2003, which implies that much of the sampling efficiency improvement of uncertainty-based methods comes from the specific way of how BiLSTM-CRF models dependency between words. In addition, the US and EDG without validation data do not perform well in MedMentions ST19. We hypothesize that this is because the transition probabilities play an important role in this dataset, but sampling many uncertain label sequences does not help the BiLSTM model.

\subsection{Sensitivity to Number of Clusters}
We choose our clustering approach and all other hyperparameters based on the validation performance after collecting the first batch in CoNLL 2003 and MedMentions ST19, and we find the performance is not sensitive to the choices unless some crucial information is missing (e.g., not considering the word shape in CoNLL 2003). 
We report the performances with varying number of clusters in each layer of our hierarchical clustering in Table~\ref{tb:hyper_stability}. We see that the micro-F1 scores are very close to each other, except in the testing set of NCBI disease dataset. We suspect the score variation in NCBI diease mainly comes from the randomness in the training process of neural networks because we conduct only one trial of experiments when filling Table~\ref{tb:hyper_stability}. The results suggest that number of clusters is a trade-off in EDG. Increasing the number of clusters decreases the bias, but increases the variance in the error decay estimation.

\begin{table*}[t!]
\centering
\begin{tabular}{|l|cc|cc|cc|}
 \hline 
    &  \multicolumn{2}{c|}{CoNLL 2003} & \multicolumn{2}{c|}{NCBI disease} & \multicolumn{2}{c|}{MedMentions ST19} \\
      & Val & Test  & Val & Test & Val  & Test      \\ \hline
EDG (J=5) & 93.0 & 88.6 &79.7 &\textbf{78.5} & 45.4 & 44.9    \\
EDG (J=10)   & \textbf{93.1} & \textbf{88.7} & \textbf{80.1} & 78.4 &  \textbf{46.0} & \textbf{45.7}    \\
EDG (J=15)  & \textbf{93.1} & 88.6 & 79.8 & 78.3 & \textbf{46.0} & \textbf{45.7} \\ \hline
EDG\_ext1 (w/o Val) (J=5) & 93.0 & 88.5 & 79.7 & 78.5 & 45.3 & 44.7  \\
EDG\_ext1 (w/o Val) (J=10) & \textbf{93.1} & \textbf{88.6} & \textbf{80.0} & 78.1 & \textbf{46.0} & \textbf{45.7}  \\
EDG\_ext1 (w/o Val) (J=15)  &  93.0 & \textbf{88.6} & 79.7 & \textbf{78.7} & 45.9 & \textbf{45.7} \\ \hline
\end{tabular}
\caption{Performance sensitivity to the number of clusters (J). Notice that the number of total clusters for the word and word + sentence feature is $J^2$ (e.g., 225 for J=15). The micro-F1 scores (\%) are the average over all the training set sizes in Figure~\ref{fig:gold_simulation}. The range of training set sizes are 40,000--200,000, 40,000--130,000, 40,000--250,000 for CoNLL 2003, NCBI disease, and MedMentions ST19, respectively. The highest F1 scores using different numbers of clusters for each sampling method are highlighted.}
\label{tb:hyper_stability}
\end{table*}



\subsection{Error Decay of Uncertainty Sampling}
\label{sec:US_analysis}
\begin{figure}[t!]
\centering
\includegraphics[width=0.85\linewidth]{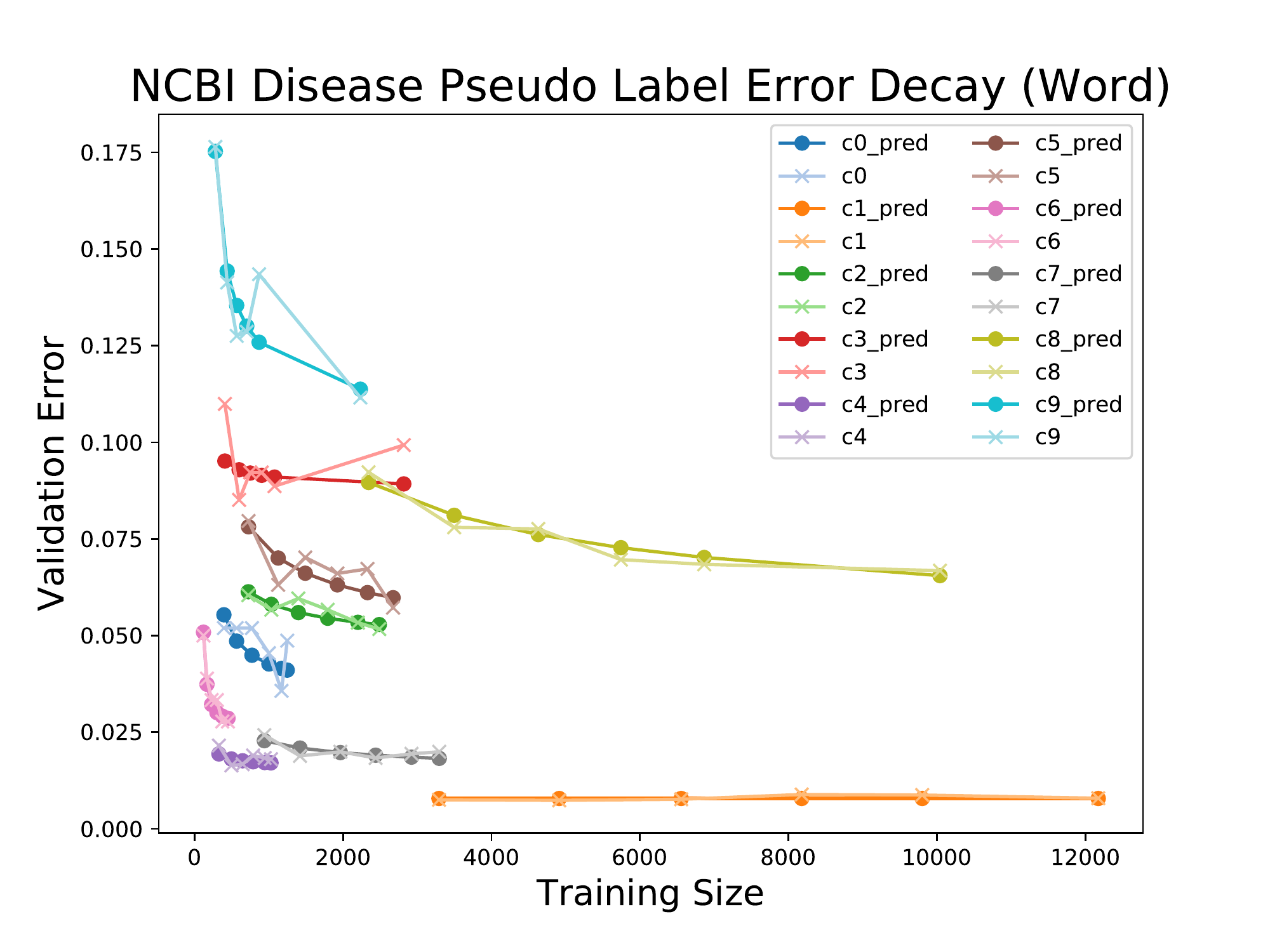}
\caption{Error decay curves of taggers trained using pseudo labels in NCBI disease dataset. The six points in each curve come from the taggers trained by 10,000, 15,000, 20,000, 25,000, 30,000, and 40,000 words. The first 30,000 words are selected randomly and uncertainty sampling (US) selects 30,000 -- 40,000 words. As in Figure~\ref{fig:error_curves}, the \textbf{x} markers on the curves are the real error and $\bullet$ means prediction from the fitting curve. The groups are formed by clustering their word embeddings, and the index of each group is presented.}
\label{fig:NCBI_US}
\end{figure}

The error curve modeling can be used not only to select the next batch but also to analyze the existing sampling strategy. For instance, Figure~\ref{fig:NCBI_US} shows that the last points in the word group 3, 8, and 9 are farther away from the fifth points in x-axis compared to the corresponding distances in other groups. This implies that uncertainty sampling tends to select samples with high errors when it chooses the 30,000th to 40,000th tokens as shown in Figure~\ref{fig:illustration}. However, the high errors do not necessarily lead to high error reduction in this dataset. This explains why US only achieves $58.3$ validation micro-F1 in Table~\ref{tb:pseudo_exp}, which is significantly worse than other methods like EDG or US+Div.

\begin{table*}[t!]
\centering
\scalebox{0.65}{
\begin{tabular}{|c|ccc|ccc|ccc|ccc|c|}
\hline
& \multicolumn{6}{c|}{Whole abstract} & \multicolumn{6}{c|}{Sentence} & \multirow{4}{*}{Avg} \\ \cline{2-13}
& \multicolumn{3}{c|}{\multirow{2}{*}{CNN for CNN}} & \multicolumn{3}{c|}{CNN for} & \multicolumn{3}{c|}{\multirow{2}{*}{CNN for CNN}} & \multicolumn{3}{c|}{CNN for}&  \\
& \multicolumn{3}{c|}{} & \multicolumn{3}{c|}{BiLSTM-CRF} & \multicolumn{3}{c|}{} & \multicolumn{3}{c|}{BiLSTM-CRF} &                 \\  \cline{2-13}
& Val& Test& Pseudo&Val& Test& Pseudo&Val& Test& Pseudo&Val& Test& Pseudo &         \\ \hline
EDG& \textbf{61.6}&	\textbf{61.9}&	\textbf{55.7}&	\textbf{68.7}&	\textbf{68.2}&	\textbf{59.2}&	\textbf{62.1}&	\textbf{60.2}&	\textbf{56.1}&	\textbf{70.0}&	68.2&	\textbf{59.8}& \textbf{62.4}\\
EDG\_ext1 (w/o Val) & 60.4&	61.5&	55.4&	67.4&	67.0&	58.6&	61.9&	60.0&	55.2&	69.0&	\textbf{69.2}&	59.1& 62.0 \\
RND& 56.5&	57.9&	51.3&	64.9&	64.1&	55.7&	58.8&	57.2&	54.1&	65.4&	65.2&	56.5& 59.0\\
Div & 57.9&	57.8&	54.2&	64.4&	64.6&	57.2&	59.3&	58.4&	53.8&	66.2&	65.7&	57.0& 59.7\\ \hline
US& 60.9&	60.9&	54.9&	68.1&	68.5&	58.3&	60.0&	59.6&	54.0&	69.7&	68.5&	59.9 &61.9\\
US+Div& 61.1&	60.4&	56.4&	65.4&	66.8&	58.3&	60.5&	58.9&	55.4&	70.1&	68.1&	59.3& 61.7 \\
US+Div+EDG\_ext2& \textbf{61.9} & 60.1 & 56.8 & \textbf{71.7} &66.1 &59.1 & 63.1 & 60.8 & \textbf{56.8} & 70.6 & 68.6 & \textbf{60.1} & 63.0 \\
BALD & 61.0 &\textbf{61.4} & 56.4 & 68.9 & \textbf{69.4} & 58.8 & 61.0 & 59.7 & 55.1 & 70.9 & 67.8 & \textbf{60.1} & 62.5 \\
BALD+EDG\_ext2& 61.3 & 60.0 & \textbf{56.9} & 69.4 & 66.5 & \textbf{59.5} & \textbf{65.0} & \textbf{62.2} & 55.9 & \textbf{72.4} & \textbf{68.7} & 59.7 & \textbf{63.1} \\
 \hline
\end{tabular}
}
\caption{The experimental setup is the same as Table~\ref{tb:pseudo_exp} except that the performances are the maximal micro-F1 (\%) of five neural networks rather than their average.}
\label{tb:pseudo_exp_max}
\end{table*}

\subsection{Weighted Class Evaluation}
\label{sec:weighted_class}
To test the controllability of our methods, we set up a variation of MedMentions ST19 where we give different penalties to different classes. Specifically, we focus on four related types: \emph{research activity,  health care activity, population group,} and \emph{spatial concept}, and assign 0.9 weights to these classes. Other 15 classes are assigned 0.1 weights. That is, when we compute micro-F1, we weight the total number of correct predictions, the total number of predictions, and the total number of entities according to the importance of classes. 

To incorporate the weights into our method, we modify $E^p_j(C_{T_t},D_V)$ in~\eqref{eq:error_opt} as

\vspace{-3mm}
\small
\begin{align}\label{eq:error_weights}
&E^p_j(C_{T_t},D_V) = \sum_{s_i \in D_V} P(g^p_j|s_i) \sum_l \left(\frac{r(y_{il})+r(\hat{y}^{C_{T_t}}_{il})}{2}\right) \mathbbm{1}( y_{il} \neq \hat{y}^{C_{T_t}}_{il}),
\end{align}
\normalsize
where $r(y_{il})$ and $r(\hat{y}^{C_{T_t}}_{il})$ are the weights of the ground truth class and the predicted class, respectively.
After this simple modification, our method boosts the weighted micro-F1 in the testing data from $34.3$ (using same $r(y)$ for all the classes) to $35.5$ (using different $r(y)$ for different classes)
 after collecting the first batch in MedMentions ST19 where the score of random sampling is $31.9$. The scores come from averaging the results of five randomly initialized CNNs.



\subsection{Results Statistics}
The micro-F1 in Figure~\ref{fig:gold_simulation} is the average of three trials. Different trials use different random initializations for the CNN. 
The performance variance among different trials is usually small.
The average standard error in the validation set across all batches and all the methods is 0.13 for synthetic data, 0.06 for CoNLL 2003, 0.28 for NCBI disease, and 0.16 for MedMentions ST19. The average standard error in the testing set is 0.18 for synthetic data, 0.14 for CoNLL 2003, 0.49 for NCBI disease, and 0.17 for MedMentions ST19.

In Table~\ref{tb:pseudo_exp}, we show the average micro-F1 of five CNNs with different initializations. Sometimes, we care more about the maximal performance, so we also report the highest F1 score out of five runs in Table~\ref{tb:pseudo_exp_max}; the results show a similar trend. When applying  two-sample t-test to the comparison of average performance in Table~\ref{tb:pseudo_exp}, we assume that every micro-F1 score is a true hidden value plus  Gaussian noise, and the variance of the noise is the same given a sampling method. Based on the assumption, the one-tailed two-sample t-test gives us $p<0.00003$ for the difference between BALD+EDG\_ext2 and BALD, between US+Div+EDG\_ext2 and US+Div, and between EDG\_ext1 and Div.


\section{Proof of Proposition 1}
\label{sec:prop1_proof}
We would like to prove that $H^p(T) = - \sum_j \hat{E}(g^p_j,C_{T}) m(g^p_j,D_A)$ is submodular and non-decreasing by assuming $\frac{d \hat{E}(g^p_j,C_{T})}{d \, m(g^p_j,T)} \leq 0$ and $\frac{d^2 \hat{E}(g^p_j,C_{T}) }{d^2 \, m(g^p_j,T)}\geq 0$ for every group $g^p_j$ in partition $p$. 

First, we prove that $-\hat{E}(g^p_j,C_{T})$ is submodular and non-decreasing. Assuming we have two subsets $X$, $Z$, and one sample $s_i$ such that $X \subseteq Z  \subseteq D_A$ and $s_i \in D_A \setminus Z$. Based on the assumption, we get $m(g^p_j,X) \leq m(g^p_j,Z)$ and $m(g^p_j,X \cup \{s_i\}) \leq m(g^p_j,Z \cup \{s_i\})$ for all $g^p_j$. Since $\frac{d \hat{E}(g^p_j,C_{T})}{d \, m(g^p_j,T)} \leq 0$, $-\hat{E}(g^p_j,C_{T})$ is non-decreasing. In order to consider the case that $m(g^p_j,X \cup \{s_i\}) \geq m(g^p_j,Z)$, we first decompose 

\footnotesize
\begin{align}\label{eq-1}
& (-\hat{E}(g^p_j,C_{X \cup \{s_i\}}))-(-\hat{E}(g^p_j,C_X)) \nonumber \\
= &(-\hat{E}(g^p_j,C_{X \cup \{s_i\}})) -(-\max(\hat{E}(g^p_j,C_{X \cup \{s_i\}}),\hat{E}(g^p_j,C_{Z})) ) \nonumber \\
 & + (-\max(\hat{E}(g^p_j,C_{X \cup \{s_i\}}),\hat{E}(g^p_j,C_{Z})) )  -(-\hat{E}(g^p_j,C_X)),
\end{align}
\normalsize
and 
\footnotesize
\begin{align}\label{eq-2}
&(-\hat{E}(g^p_j,C_{Z \cup \{s_i\}}))-(-\hat{E}(g^p_j,C_Z)) \nonumber \\
= & (-\hat{E}(g^p_j,C_{Z \cup \{s_i\}})) -(-\min(\hat{E}(g^p_j,C_{X \cup \{s_i\}}),\hat{E}(g^p_j,C_{Z})) ) \nonumber \\
 & + (-\min(\hat{E}(g^p_j,C_{X \cup \{s_i\}}),\hat{E}(g^p_j,C_{Z})) ) -(-\hat{E}(g^p_j,C_Z)).
\end{align}
\normalsize
Based on mean value theorem and 
\footnotesize
\begin{align}\label{eq-3}
\small
& (-\hat{E}(g^p_j,C_{X \cup \{s_i\}}))  -(-\max(\hat{E}(g^p_j,C_{X \cup \{s_i\}}),\hat{E}(g^p_j,C_{Z})) ) \nonumber \\
=& (-\min(\hat{E}(g^p_j,C_{X \cup \{s_i\}}),\hat{E}(g^p_j,C_{Z})) ) -(-\hat{E}(g^p_j,C_Z)),
\end{align}
\normalsize
we get
\footnotesize
\begin{align}\label{eq-4}
&(-\hat{E}(g^p_j,C_{X \cup \{s_i\}}))-(-\hat{E}(g^p_j,C_X)) - \nonumber \\
 & (-\hat{E}(g^p_j,C_{Z \cup \{s_i\}}))+(-\hat{E}(g^p_j,C_Z)) \nonumber \\
= & (-\max(\hat{E}(g^p_j,C_{X \cup \{s_i\}}),\hat{E}(g^p_j,C_{Z})) ) \nonumber \\
& -(-\hat{E}(g^p_j,C_X)) - (-\hat{E}(g^p_j,C_{Z \cup \{s_i\}})) \nonumber \\
& +(-\min(\hat{E}(g^p_j,C_{X \cup \{s_i\}}),\hat{E}(g^p_j,C_{Z})) )  \nonumber \\
=& (-\frac{d \hat{E}(g^p_j,C_X)}{d \, m(g^p_j,X)}|_{m(g^p_j,X)=x_1} + \frac{d \hat{E}(g^p_j,C_X)}{d \, m(g^p_j,X)}|_{m(g^p_j,X)=x_2}) \cdot \nonumber \\
 & \min(m(g^p_j,Z)-m(g^p_j,X), |s_i|), 
\end{align}
\normalsize
where $m(g^p_j,X)< x_1 < \min(m(g^p_j,X \cup \{s_i\}),m(g^p_j,Z))$ and $\max(m(g^p_j,X \cup \{s_i\}),m(g^p_j,Z))< x_2 < m(g^p_j,Z \cup \{s_i\})$.

Since $x_2 > x_1$ and $\frac{d^2 \hat{E}(g^p_j,C_X)}{d^2 \, m(g^p_j,X)}\geq 0$, we know \\ 
$\frac{d \hat{E}(g^p_j,C_X)}{d \, m(g^p_j,X)}|_{m(g^p_j,X)=x_1} \leq \frac{d \hat{E}(g^p_j,C_X)}{d \, m(g^p_j,X)}|_{m(g^p_j,X)=x_2}$, which leads to $(-\hat{E}(g^p_j,C_{X \cup \{s_i\}}))-(-\hat{E}(g^p_j,C_X)) \geq (-\hat{E}(g^p_j,C_{Z \cup \{s_i\}}))-(-\hat{E}(g^p_j,C_Z))$, so $-\hat{E}(g^p_j,C_T)$ is submodular and non-decreasing.

Finally, $H^p(T) = - \sum_j \hat{E}(g^p_j,C_{T}) m(g^p_j,D_A)$ is submodular and non-decreasing because $m(g^p_j,D_A)$ does not change over $X$, and the linear combination of submodular and non-decreasing functions with non-negative weights is still submodular and non-decreasing.








\section{Implementation Details}
\label{sec:implement_details}
In Equation~\eqref{eq:UD}, the difference between the current training data size $t_m$ and previous training data size $t_p$ is always 20,000 words for real-world datasets, and 2,000 words for the synthetic dataset in our experiments. When choosing samples using Equation~\eqref{eq:UD}, we need to be careful about the starvation problem. That is, some types of samples are not selected in the recent history, and the samples would have low uncertainty changes which further prevents them from being selected in the future. To mitigate this issue, we alternate between using the scores in Equation~\eqref{eq:UD} and the current uncertainty $u_i^{t_m}$ to choose the next batch. For example, when plotting the performance of BALD+EDG\_ext2 in Figure~\ref{fig:gold_simulation}, we select the first batch using BALD+EDG\_ext2 (i.e., Equation~\eqref{eq:UD}) and the second batch using only BALD (i.e.,  $u_i^{t_m}$), the third batch using BALD+EDG\_ext2, and so on. The same strategy is applied to US+Div+EDG\_ext2 as well.

In~\eqref{eq:error_opt}, we find that $w_j$ and $v_{tj}$ could be set as 1 in most of the cases. However, we observe some predicted error decay curves collapse into a flat line (i.e., $b_j=0$ in Equation~\eqref{eq:approx_error}) due to the unstable performance in validation set. To increase robustness, we set $w_j=\min(100,m(g^p_j,D_V))$, and $v_{tj} = 3$ if $t=\arg\min_{x} E^p_j(C_{T_x},D_V)$ (i.e., lowest error for $j$th group across $t$) and $v_{tj} = 1$ otherwise in our experiments, and optimize~\eqref{eq:error_opt}
using Newton Conjugate-Gradient~\citep{nash1984newton}. 

In~\eqref{eq:selection_features}, we use geometric mean to combine multiple features because we usually want a sample that has large error reduction in all the groups it belongs to. Our preliminary experiments indicate that using geometric mean is better than arithmetic mean. The smoothness constant $\epsilon$ in~\eqref{eq:selection_features} should be proportional to the size of the dataset $D_A$ because larger error reduction could be made in a larger dataset. In our experiments, we set $\epsilon$ to be $0.01$ for MedMentions ST19, $0.001$ for NCBI disease and CoNLL 2003 dataset.







\end{document}